\documentclass[acmsmall]{acmart}

\AtBeginDocument{%
  }

\setcopyright{acmcopyright}
\copyrightyear{2023}
\acmYear{2023}
\acmDOI{XXXXXXX.XXXXXXX}

\acmJournal{TOMM}
\acmVolume{00}
\acmNumber{0}
\acmArticle{000}
\acmMonth{1}

\usepackage[ruled,noline,linesnumbered]{algorithm2e}
\usepackage{multirow}
\usepackage[labelformat=simple]{subcaption}

\usepackage{appendix}

\newtheorem{myDef}{Definition}
\begin{document}

\title{\texorpdfstring{CD$^2$}{CD2}: Fine-grained 3D Mesh Reconstruction With Twice Chamfer Distance}

\author{Rongfei Zeng}
\authornote{Both authors contributed equally to this paper.}
\authornotemark[2]
\email{zengrf@swc.neu.edu.cn}

\author{Mai Su}
\authornotemark[1]
\email{sumai1998@foxmail.com}

\author{Ruiyun Yu}
\email{yury@mail.neu.edu.cn}

\author{Xingwei Wang}
\authornote{Corresponding author}
\email{wangxw@mail.neu.edu.cn}
\affiliation{%
  \institution{Northeastern University}
  \streetaddress{No. 11, Wenhua Street, Heping District}
  \city{Shenyang}
  \state{Liaoning}
  \country{China}
  \postcode{110819}
}

\renewcommand{\shortauthors}{Zeng, et al.}

\begin{abstract}
Monocular 3D reconstruction is to reconstruct the shape of object and its other information from a single RGB image. In 3D reconstruction, polygon mesh, with detailed surface information and low computational cost, is the most prevalent expression form obtained from deep learning models. However, the state-of-the-art schemes fail to directly generate well-structured meshes, and we identify that most meshes have severe Vertices Clustering (VC) and Illegal Twist (IT) problems. By analyzing the mesh deformation process, we pinpoint that the inappropriate usage of Chamfer Distance (CD) loss is a root cause of VC and IT problems in deep learning model. In this paper, we initially demonstrate these two problems induced by CD loss with visual examples and quantitative analyses. Then, we propose a fine-grained reconstruction method CD$^2$ by employing Chamfer distance twice to perform a plausible and adaptive deformation. Extensive experiments on two 3D datasets and comparisons with five latest schemes demonstrate that our CD$^2$ directly generates a well-structured mesh and outperforms others in terms of several quantitative metrics.
\end{abstract}



\maketitle

\section{Introduction}
Monocular 3D mesh reconstruction, boosted by impressive deep learning technology, is a fundamental and fascinating topic in the community of computer vision. It aims to generate detailed 3D information of object's surface, orientation, etc. in mesh format from a single 2D image \cite{mescheder_occupancy_2019}. The prevalent data format of mesh, which consists of hundreds of vertices and faces, has plenty of benefits and advantages. For instance, mesh can efficiently and accurately capture the details of 3D objects' surface and describe almost every shape in the world, such as cars, boats, and airplanes in ShapeNet dataset \cite{chang2015shapenet}. Meanwhile, the quality of mesh has been drastically improved by the promising deep learning technology and large-scale datasets in recent years \cite{sun_pix3d:_2018,chang2015shapenet}. All these merits empower explicit 3D mesh reconstruction with deep learning \cite{ZengIncentive} to have widespread applications in 3D printing, automatic drive, virtual reality, augmented reality, medical diagnosis, and online shopping \cite{nguyen_graphx-convolution_2019}. 

In the deep-learning-enabled mesh reconstruction, Chamfer Distance (CD) \cite{barrow1977parametric} is a universal and paramount component used as a loss function of deep learning model as well as a metric to evaluate the quality of generated 3D mesh. CD calculates the average of pair-wise nearest neighbour distance between a synthetic mesh and a ground truth object. CD is preferable for explicit mesh reconstruction due to its efficient computation and flexibility with different volumes of points. The vast majority of impressive works such as Atlasnet \cite{groueix_papier-mache_2018}, TMN \cite{pan_deep_2019}, and Pixel2Mesh \cite{wang2018pixel2mesh} employ CD as a loss function or part of loss function for model training and achieve desirable reconstruction performance. 

\begin{figure}[!t]
\centering
\includegraphics[scale=0.12]{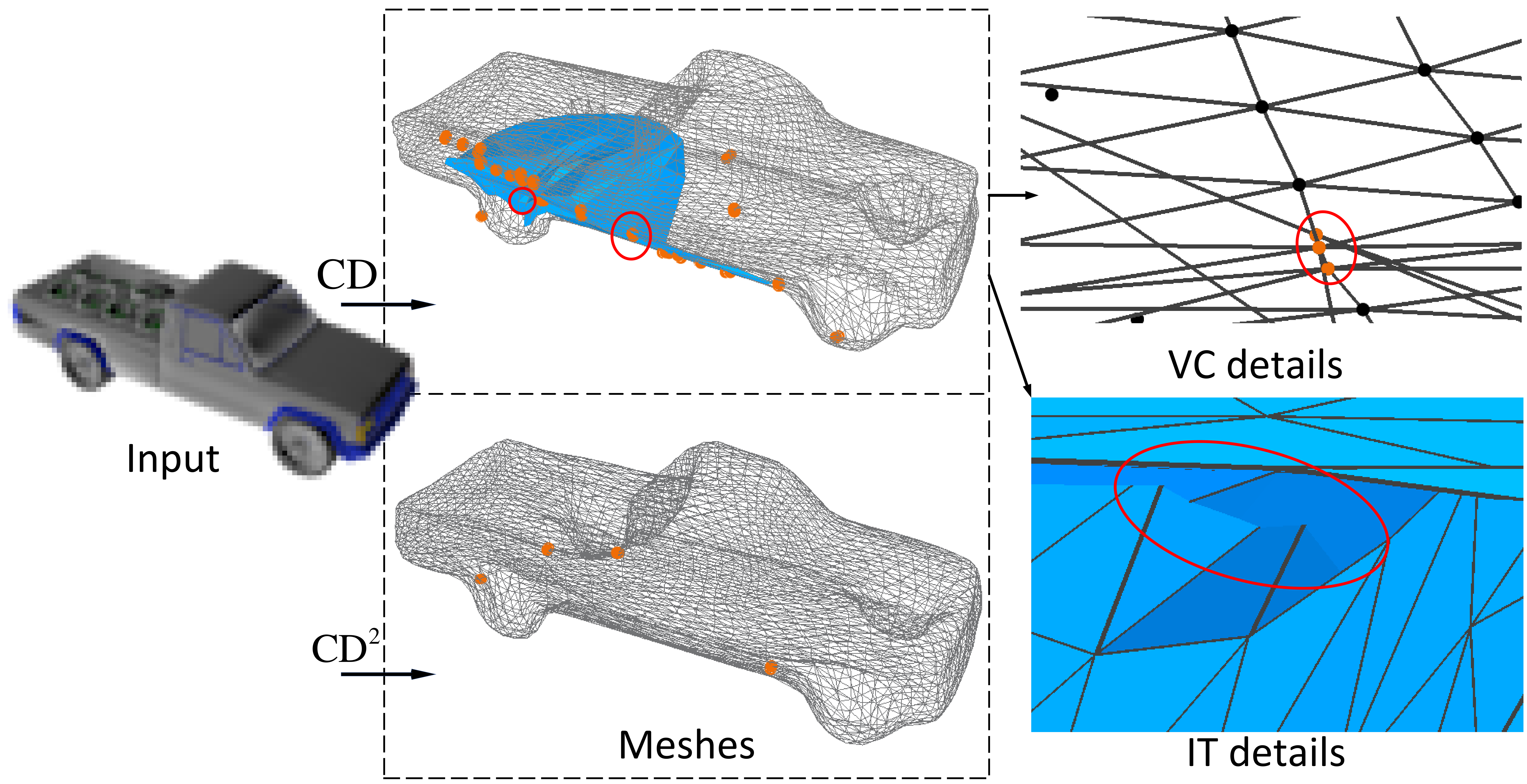}
\caption{IT and VC problems in 3D mesh reconstructions and our improvement of CD$^2$.}
\label{cdproblem}
\end{figure}

However, CD has some intrinsic deficiencies and previous studies have been published to identify them. For instance, Li et al. find that CD may suffer from the local optimum problem in its nearest neighbor search \cite{li_lbs_2019}. Wu et al. reiterate that CD is insensitive to point density distribution and also is prone to be impacted by outliers \cite{wu2021density}. Jin et al. demonstrate that CD may not be faithful visually and structurally \cite{jin2020dr}. Achlioptas et al. show that CD is inclined to generate overcrowded points in some visible areas \cite{achlioptas2018learning}. Wagner et al. empirically prove that directly optimizing CD will ignore the details of some structures \cite{wagner_neuralqaad:_2022}. Meanwhile, researchers also propose some variant CDs, such as a structured CD \cite{li_lbs_2019}, a sharper version of CD \cite{lim_convolutional_2019}, an adaptive CD \cite{wang2020deep}, a probabilistic CD \cite{li_usip:_2019}, an augmented CD \cite{chen_deep_2020}, a density-aware CD \cite{wu2021density}, etc., and attempt to fix the aforementioned defects caused by CD loss.

Besides that, we find that CD loss still suffers from IT and VC problems, which have not been discovered before. As shown in Fig. \ref{cdproblem}, IT problem refers to the phenomenon that some faces locally and irrationally intersect with others or interpose into the inner regions, while VC problem describes the scenario where many vertices cluster around a single ground truth point. Empirically, both IT and VC problems are attributed to CD loss function which offers incorrect deformation directions in the brute-force nearest neighbour search. The consequence of these two problems is performance degradation caused by wasting limited vertices in some invisible and overcrowded regions and distorting the mesh into an unusable structure. Furthermore, these severe problems might malfunction the downstream applications of 3D reconstruction, such as 3D printing, rendering, etc. In a nutshell, IT and VC problems introduced by CD loss should be considered seriously in the explicit mesh reconstruction. 

In this paper, we take a deep dive into the IT and VC problems and endeavor to improve the quality of reconstructed mesh with the idea of twice CD calculation in deep learning method. Staring from the identification of IT and VC problems in both visual and quantitative ways, we find that some vertices of generated mesh move in incorrect directions or at an inappropriate speed. Following these observations, we present our proposal CD$^2$ to perform a fine-grained mesh deformation process. The proposed CD$^2$ calculates the CD loss twice and excludes some excessively-moving vertices from further deformation with some criteria, which makes vertices deform with a comparable speed and correct directions. Comparing with five impressive works on two datasets ShapeNet and Pix3D, we demonstrate that our scheme CD$^2$ can effectively mitigate IT and VC problems and yield a well-structured 3D mesh model.

The contributions of this paper are threefold: 
\begin{itemize}
    \item To the best of our knowledge, we are the first to identify IT and VC problems induced by CD loss in deep learning model of 3D reconstruction and also provide both visual examples and quantitative analyses of these two problems. Statistical results show that 28\% vertices have VC problems in impressive Atlasnet \cite{groueix_papier-mache_2018} when $\rho=0.5$, and 28.1\% faces have IT problems in Total3D \cite{nie_total3dunderstanding:_2020}.
    
    \item We propose an innovative approach CD$^2$ with twice CD calculation to mitigate IT and VC problems by moving vertices at a moderate speed and in rational directions. Our proposal CD$^2$ can achieve a fine-grained deformation process.
    
    \item Visual demonstrations show our CD$^2$ generates more plausible and well-structured meshes than other five baseline schemes. Meanwhile, quantitative evaluation results reveal that our CD$^2$ outperforms other baselines in terms of several new metrics. These quantitative metrics provided by our work consider the distance or mapping relation of mesh and can achieve a faithful and appropriate measuring capability.
\end{itemize}

The remainder of this paper is organized as follows. In Section \ref{sec:problem_state}, we identify the IT and VC problems and formulate these two problems. Section \ref{sec:method} presents our proposed method CD$^2$. Extensive experiment results are presented in Section \ref{sec:experiment}, followed by related work in Section \ref{sec:relatedwork}. Section \ref{sec:conclusion} concludes the entire paper.

\section{Problem Statement And Analysis} 
\label{sec:problem_state}
\subsection{Chamfer Distance}
Chamfer distance, proposed by Barrow et al. in \cite{barrow1977parametric}, is a commonly-used metric to measure the average of pair-wise nearest distance between two point sets. In deep learning model of 3D mesh reconstruction, CD is chosen as the loss function in the optimization process and the evaluation metric of 3D mesh quality. Mathematically, CD is defined as

\begin{equation}
{d_{C\!D}}({S_1},\!{S_2}) = {d_{C\!D1}}({S_1},\!{S_2})+ {d_{C\!D2}}({S_1},\!{S_2}) =
\frac{1}{|S_1|} \! \sum\limits_{x \in {S_1}} \! {\mathop {\min }\limits_{y \in {S_2}} } ||x - y||_2^2 + \frac{1}{|S_2|} \! \sum\limits_{y \in {S_2}} \! {\mathop {\min }\limits_{x \in {S_1}} } ||x - y||_2^2,
\label{eq:CD}
\end{equation} 

where $S_1 \subseteq {\mathbf{R}^3}$ is the ground truth point set and $S_2 \subseteq {\mathbf{R}^3}$ is the reconstructed mesh vertex set\footnote{We use the term vertex to denote point in the reconstructed mesh for readability.}. CD is composed of two parts $d_{C\!D1}({S_1},\!{S_2})$ and $d_{C\!D2}({S_1},\!{S_2})$. The term $d_{C\!D1}({S_1},\!{S_2})$ computes the average distance from a point in ${S_1}$ to its nearest vertex in ${S_2}$, while the second term ${d_{CD2}}({S_1},{S_2})$ evaluates the average distance from a vertex in mesh to its nearest point in ${S_1}$.

\begin{table}[!t]
\caption{Notations frequently used in this paper.}
\setlength{\tabcolsep}{5.2mm}
\begin{tabular}{ll}
\toprule
\textbf{Notations} & \textbf{Explanations}     \\
\midrule
${S_1}$ & The ground truth point set of a target 3D object \\
${S_2}$ & The vertex set of synthetic mesh\\ 
${V_i}$ & The $i$th vertex in the mesh set ${S_2}$\\ 
${P_j}$ & The $j$th point in the ground truth point set ${S_1}$\\ 
${\phi{(V_i)}}$ & ${V_i}$'s nearest point in ${S_1}$ \\
${\psi{(P_j)}}$ & ${P_j}$'s nearest vertex in ${S_2}$ \\ 
${\mathbb{P}_{{V_i}}}$ & Points in ${S_1}$ whose nearest vertex is ${V_i} \in S_2$ \\ 
${\mathbb{V}_{{P_j}}}$ & Vertices in ${S_2}$ whose nearest point is ${P_j} \in S_1$ \\
\bottomrule
\end{tabular}
\label{tab:notations}

\end{table}

In practice, we have ${\rm{|}}{S_1}| \gg |{S_2}|$. For instance, each 3D object consists of 10,000 ground truth points in Pix3D dataset \cite{nie_total3dunderstanding:_2020} and 30,000 points in ShapeNet \cite{chang2015shapenet}, while the sphere template for deformation only use 2,562 vertices in both Total3D \cite{nie_total3dunderstanding:_2020} and Atlanset \cite{groueix_papier-mache_2018} due to limitations of computing resources. This constraint implies that we should not waste limited vertices in some valueless inner regions and vertices should be used efficiently to model the surface of a target object. For simplicity, we list some commonly-used notations in Table~\ref{tab:notations}.

\subsection{A Toy Example of Mesh Deformation Process}
\begin{figure}[!t]
\begin{minipage}[b]{\linewidth}
    \centering
    \subfloat[A toy chair with CD deformation.]{\label{toy-cd}\includegraphics[scale=0.2]{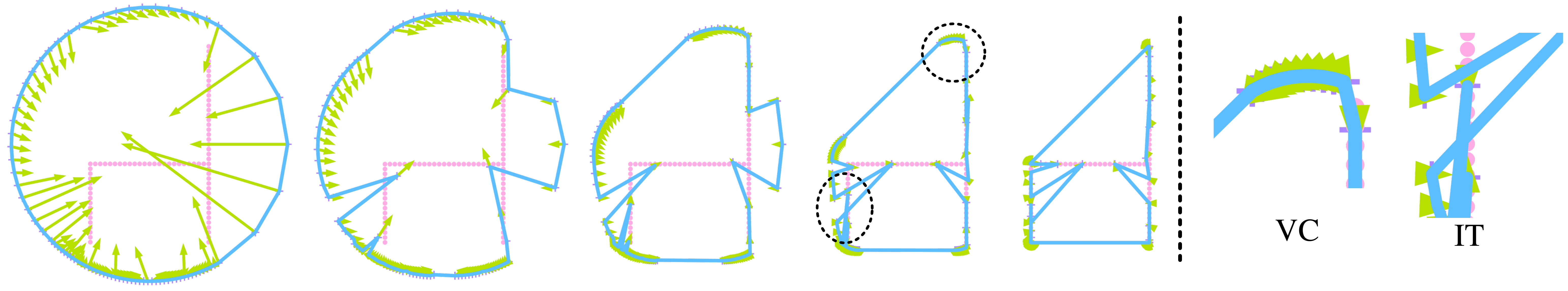}}
\end{minipage} 
\begin{minipage}[b]{\linewidth}
    \centering
    \subfloat[A toy chair with CD$^2$ deformation.]{\label{toy-cd2}\includegraphics[scale=0.2]{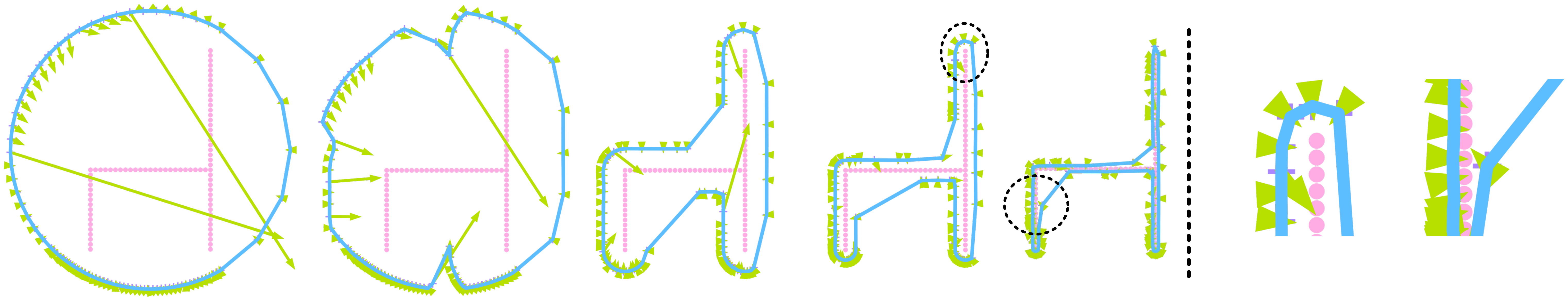}}

\end{minipage}

\caption{The deformation example of a toy 2D chair and the details of VC and IT problems. We use pink dots to denote ground truth points, purple pluses to denote mesh vertices, blue lines to denote edges between vertices, and cyan arrows to denote deformation directions of vertices. We also zoom in on the IT and VC problems in the right subfigure.}
\label{figToyexample}
\end{figure}

In this subsection, we present a toy example to demonstrate the mesh deformation process with CD loss in Fig. \ref{toy-cd}. In this example, we use 81 pink dots (ground truth set $S_1$) to model a 2D wooden chair and 80 purple pluses (vertex set $S_2$) connected by blue edges to represent a mesh template. We first calculate CD between these two sets and use the API $torch.Tensor.backward()$ in PyTorch to obtain the gradient of each vertex. The deformation directions are shown by cyan lines with arrow. During each deformation, we move all the vertices along their deformation directions which are the reverse directions of their gradients. This deformation process iterates until the CD loss is minimized or the total number of iterations exceeds the predefined threshold. When the deformation process is finished, the generated mesh imperfectly models the target chair. Some deformation snapshots are given in Fig. \ref{toy-cd}.

We have two observations from this deformation process in Fig. \ref{toy-cd}. (1) Many vertices in $S_2$ are located in small and specific regions like the end of legs and chair back in this example. We explicitly present these overcrowded vertices with dotted circles and name this phenomenon as Vertices Clustering (VC). Actually, there is no need to waste the excessive number of vertices to model chair back and the end of legs, and some vertices are preferred to approach to other important details like the cushion surface of chair. (2) In Fig. \ref{toy-cd}, there exist some unnecessary and intersected lines shown in the dotted circle of the 4th subfigure. We zoom in on this part in the 6th subfigure. Similarly, some vertices rush into the interior of chair leg and generate the unexpected twists of faces in 3D deformation process. Few vertex is needed to model the interior of a chair, and it is the best practice to use enough vertices to describe the surface of cushion. The aggressive deformation of these vertices causes the generated mesh to have intersected and twisted faces and not properly align with the target object. We call this phenomenon Illogical Twist (IT). Both VC and IT are induced by CD loss which only considers the point-level similarity in the brute-force nearest neighbour search, ignoring the edge relation of mesh and the overall shape of target object.

In Fig. \ref{toy-cd2}, we also provide an example to demonstrate the performance improvement of our scheme CD$^2$. Compared with the 5th subfigure of Fig. \ref{toy-cd}, our scheme CD$^2$ generates a well-structured 2D chair which can model the target object perfectly. For the VC problem, the 6th subfigure of Fig. \ref{toy-cd2} shows that our CD$^2$ moves vertices towards different ground truth points instead of a single point. In other words, CD$^2$ drastically relieves VC problems. Moreover, IT problem disappears in our CD$^2$. More results of performance improvement are given in Section \ref{sec:experiment}.

\subsection{Mathematical Definitions of VC and IT Problems}
\label{sec:DefofVCIT}

\textbf{Vertices Clustering.} 
VC describes a phenomenon that multiple mesh vertices overcrowd or overlap with each other in a small region as shown in Fig. \ref{cdproblem}. The formal definition of VC is given as follows: 
\begin{myDef} Vertices Clustering: Let $\bar d({S_1})$ be the average nearest neighbor distance between points in $S_1$, and define the distance threshold between two vertices in $S_2$ as ${\sigma_{V\!C}} = \rho \bar d({S_1})$, where the coefficient $\rho$ is a hyper-parameter. Then, we have two criteria to evaluate the VC problem: (1) ${N_{VC}}$, the size of VC vertex set $S_{V\!C} \subseteq {S_2}$, where each vertex $V_i \in S_{V\!C}$ has the nearest neighbor $V_j \in S_{V\!C}$ and this nearest neighbor distance is smaller than ${\sigma_{V\!C}}$, i.e., $d({V_i},{V_j}) < {\sigma _{VC}}$. In other words, we can use ${N_{V\!C}}= |S_{V\!C}|$ to evaluate the VC problem; and (2) ${N_{VC'}}$, the number of vertex $V_i \in S_{V\!C'} (S_{V\!C'} \subset S_{V\!C})$ which has an identical nearest ground truth point ${P_l} \in S_1$ with other $V_j \in S_{V\!C'}$, i.e., $(d({V_i},{V_j}) < {\sigma _{VC}}) \& (\phi ({V_i}) = \phi ({V_j})={P_l})$.
\label{def:vc}
\end{myDef}
We have some discussions about this definition. Initially, we argue that the VC metric endeavours to find vertices which are extremely close to each other during deformation. In practice, it is desirable to evenly distribute vertices over the surface of the target object. For this goal, we can coordinate the deformation process with this VC constraint to alleviate the overcrowding problem and efficiently distribute vertices in a fine-grained manner. In addition, these two criteria of VC can be used as the quality evaluation metric of synthetic mesh as well. Finally, we can easily find that the second criterion is much stricter than the first one. 

According to our formal definition, we find that VC problem exists universally in previous CD-based schemes.
Taking a rifle in Atlasnet \cite{groueix_papier-mache_2018} for example, we find that there are 1429 VC vertices of the first type (i.e., $N_{V\!C}=1429$) among 2562 total vertices and 1176 VC vertices of the second type (i.e., $N_{V\!C'}=1176$) when $\rho=0.5$. The average percentage of first-type VC vertices $(N_{V\!C}/{|S_2|})$ and the average percentage of second-type VC vertices $({N_{V\!C'}}/{|S_2|})$ are separately 55.78\% and 45.90\% for the class of rifle. The average percentages separately reach 28.1\% and 20.9\% for all the categories in ShapeNet. We also present the results of VC problem in Atlasnet for $\rho=0.25$ in Table~\ref{tab:VCandIT}. From this table, we find the classes of rifle, lamp, and airplane are prone to have VC problems. In sum, VC is a serious problem for CD loss.

\textbf{Illogical Twist.} IT refers to the situation where some parts of mesh intersect with itself or rush into the inner region as shown in Fig. \ref{cdproblem}. The IT problem can be formally defined as follows:
\begin{myDef} Illegal Twist: For two faces $f_i$ and $f_j$, let $v(f_i,f_j)$ denote the number of shared vertices between these two faces. If $v({f_i},{f_j}) = 0$, we use the algorithm in \cite{yi2008fast} to determine whether $f_i$ intersects with $f_j$ or not and then construct a IT face set $S_{IT}$ with this algorithm. Then, we have two criteria to evaluate IT problem: (1) $F_{IT}$, the size of IT face set, $F_{IT} = |S_{IT}|$; and (2) $V_{IT}$, the number of vertices which belong to the faces in $S_{IT}$. 
\label{def:it}
\end{myDef}

The IT phenomenon also occurs universally in reconstructed meshes of Total3D and Atlasnet in Fig. \ref{fig:atlas_it_VC} and Fig. \ref{figmesh}. Fig. \ref{fig:atlas_it_VC} shows this problem with blue marked faces, while Fig. \ref{figmesh} presents the IT problem in wireframes from different views. Besides visual results, we also use two types of IT metrics given in Definition \ref{def:it} to evaluate the quality of generated mesh for Atlasnet in Table~\ref{tab:VCandIT}. It can be found that meshes of table, rifle, chair, and bench have severe IT problems. For the class of table, there exist 809.5 first-type IT faces on average and 677.8 vertices of the second-type IT metric. The percentages are separately 15.8\% and 26.5\%. These results are nontrivial, and the output yields an unacceptable and valueless mesh model which is not suitable for applications in various downstream tasks.

Why do VC and IT problems occur in the deformation process? Typically, ground truth points are uniformly distributed across the 3D object's surface, but objects in the image are presented in a fixed view. Learning from images with partially visible objects and unbalanced confidence levels leads to unreasonable gradient directions and a huge gap of deformation velocity. In other words, visible parts of objects in the image have a larger deformation gradient than those invisible parts. Even worse, some excessively-deformed vertices still have a large deformation velocity after arriving at the target position and cannot stop moving in the brute-force nearest neighbour search, as shown in Fig. \ref{toy-cd}. This over-deformation ultimately results in the intersection and twist of faces or vertex clustering in a small region. From the above illustration, we can identify VC and IT vertices and then orchestrate them with an appropriate velocity to prevent the output mesh from VC and IT problems. 

\section{Our Fine-grained Mesh Reconstruction Scheme CD$^2$}
\label{sec:method}

In this section, we provide two fine-grained 3D mesh reconstruction schemes CD$^2$ both of which divide a deformation iteration into two sub-steps and separately compute the CD metric in each sub-step. Besides the initial CD calculation, the first sub-step also involves vertices exclusion according to CD losses for all the vertices and points. Then, we compute the CD metric twice for residual vertices and deform them in the second sub-step. Our fine-grained scheme CD$^2$ does not consider all the vertices equally and orchestrates the deformation process in a moderate and adaptive manner. According to different considerations in vertices exclusion of the first step, our proposal CD$^2$ involves two versions, i.e., the distance-oriented CD$^2$ and the mapping-oriented CD$^2$.

\subsection{The Distance-oriented \texorpdfstring{CD$^2$}{CD2}}
We first present our distance-oriented CD$^2$ which utilizes the distance information to exclude those aggressive vertices. In the first step of distance-oriented CD$^2$, we compute a CD metric for all the vertices, identify those excessively-deformed vertices, and then exclude them from the subsequent deformation process according to distance information. In the second step, our scheme computes another CD metric for residual vertices and moves them according to these new gradients. Before introducing the details of vertex exclusion method in the first step, we define four critical data structures used in the following paper as follows: 

\begin{itemize}
    \item $Dist_1$ and $Index_1$. In the nearest distance list $Dist_1$, the element $Dist_1[i]$ is the distance from point $P_i \in S_1$ to its corresponding nearest vertex $V_j = \psi({P_i}) \in S_2$. We also put the index of this nearest vertex $j$ into the list $Index_1$, i.e., $Index_1[i]=j$.
    \item $Dist_2$ and $Index_2$. In the nearest distance list $Dist_2$, the element $Dist_2[i]$ is the distance from vertex $V_i \in S_2$ to its corresponding nearest point $P_j =\phi({V_i}) \in S_1$. The index of this nearest point $j$ is also added to the list $Index_2$, i.e., $Index_2[i]=j$.
\end{itemize}

In the following, we provide a simple and useful method to identify those vertices which are extremely close to the ground truth points and then exclude them from further deformation. This method conforms to our empirical insight that we should move all the vertices at a comparable velocity to alleviate the VC and IT problems. In detail, we construct an exclusion vertices set $S_{2d}$ and its size is determined by two factors, i.e.,
\begin{math}
    |S_{2d}| = max \{{p_d} \times |{S_2}|, |Dis{t_2} < {d_T}|\},
\end{math}
where the threshold $p_d \in (0,1)$ is the percentage of vertices excluded from the next CD computation and $|Dis{t_2} < {d_T}| \in [0, |S_2|]$ is the number of vertices whose nearest point distance is smaller than the threshold $d_T$. With the data structures $Dist_2$ and $Index_2$, we can easily find those vertices within a certain nearest distance. Then, we can construct a new vertex set $S'_2 = S_2 - S_{2d}$. Meanwhile, we can obtain the set ${S_{1d}}$ by mapping ${S_{2d}}$ with $Index_2$ and get the corresponding new ground truth point set $S'_1$ for further CD computation. In the second step, we compute another CD loss for these new sets ${S'_1}$ and ${S'_2}$ and only move vertices in ${S'_2}$ to the ground truth points. The detailed algorithm is presented in Algorithm~\ref{algCD2_1}.

\begin{algorithm}[!t]
\caption{The Distance-oriented CD$^2$}
\label{algCD2_1}
\KwIn{${S_1}$,${S_2}$,${{{p}}_{d}}$,$d_T$}
\KwOut{${d_{\rm{C{D^2}}}}({S_1},{{\rm{S}}_2})$}
$Dist_1,Dist_2,index_1,index_2 = chamfer\_distance({S_1},{S_2})$ \par
${\rm{|}}{{{S}}_{2d}}{\rm{|}} =  {max({{{p}}_{d}} \times |S_2|, |Dis{t_2} < {d_T}|)}$ \par
        \For{$i=1$ to $|S_2|$}{
                \If{$Dis{t_2}[i]$ is top ${\rm{|}}{{{S}}_{2d}}{\rm{|}}$ smallest}{
                    Append ${S_2}[i]$ to ${S_{{2d}}}$\par
                    Append ${S_1}[Inde{x_2}[i]]$ to ${S_{{1d}}}$
                    }
                    }
        $\textit{${S'_1}$} = \textit{${S_1}$} - \textit{${S_{{1d}}}$}$\par
        $\textit{${S'_2}$} = \textit{${S_2}$} - \textit{${S_{{2d}}}$}$\par
        
        $Dis{{t'}_1}, Dis{{t'}_2}, inde{x'}_1, inde{x'}_2 = chamfer\_distance({S'_1},{S'_2})$ \par
        ${d_{\rm{C{D^2}}}}({S_1},{{\rm{S}}_2}) =  Mean(Dis{{t'}_1})+Mean(Dis{{t'}_2})$ 
\end{algorithm}

In Line 1 of Algorithm \ref{algCD2_1}, we calculate the first CD loss according to Eq. (1). In practice, a common python function called $chamfer\_distance()$\footnote{https://github.com/ThibaultGROUEIX/ChamferDistancePytorch} is adopted to obtain this loss, and this function also return four data structures defined above. It is also true for Line 11. In Line 2, $|Dis{t_2} < {d_T}|$ is the number of vertices with nearest distance to the ground truth point smaller than $d_T$. From Algorithm \ref{algCD2_1}, we can find that both mesh vertices and ground truth points vary in the second step of each deformation iteration, which makes our distance-oriented CD$^2$ more adaptive than the traditional CD. In addition, the selection of vertex for deformation enables the generated mesh to approach to the target 3D object in a synchronous manner, which further relieves VC and IT problems. Finally, we should note that the computational cost is not increased much, although we compute CD twice. Both the number of vertices and points are reduced in the second CD calculation. Our distance-oriented CD$^2$ still dominates EMD in terms of computational cost, which will be further analyzed in Section \ref{sec:time_c}.

\begin{algorithm}[t]
\caption{The Mapping-oriented CD$^2$}
\label{algCD2_2}
\KwIn{${S_1}$, ${S_2}$, $pv{i_t}$, $pv{i_p}$}
\KwOut{${d_{{\rm{CD}}^2}}({S_1},{S_2})$}
\SetKwFunction{FExclude}{Exclude}
\SetKwProg{Fn}{Function}{:}{}
\Fn{\FExclude{$Index_1$, $Index_2$, $pv{i_t}$, $pv{i_p}$, ${S_1}$, ${S_2}$}}{
${\mathbb{P}_{{V_i}}} = \emptyset ,i = 1,2,...,|{S_2}|$\par
\For{$j=1$ to $|Index_1|$}{
$jnn=Inde{{x}_{1}}[j]$\par
            ${{\mathbb{P}}_{{{V}_{jnn}}}}={{\mathbb{P}}_{{{V}_{jnn}}}}\cup {{S}_{1}}[j]$ }
        \If{${\rm{CD}}_t^2$ is applied}{
        \For{$i=1$ to $|S_2|$}{
           
            \If{$|{{\mathbb{P}}_{{{V}_{i}}}}|$ $ > $ $pv{i_t}$ }{
                    Append ${S_2}[i]$ to ${S_{{2d}}}$\par
                   }}}
                   
        \If{${\rm{CD}}_p^2$ is applied}{
        \For{$i=1$ to $|S_2|$}{
            \If{$|{{\mathbb{P}}_{{{V}_{i}}}}|$ is top $pv{i_p}$ largest}{
                    Append ${S_2}[i]$ to ${S_{{2d}}}$\par
                   } }}

        $\textit{${S'_2}$} = \textit{${S_2}$} - \textit{${S_{{2d}}}$}$\par
    \KwRet $\textit{${S'_2}$}$
}

 $Dist_1,Dist_2,index_1,index_2 = chamfer\_distance({S_1},{S_2})$ \par
 $\textit{${S'_2}$} = Exclude(Index_1, Index_2, pv{i_t}, pv{i_p}, {S_2})$ \par
 $\textit{${S'_1}$} = Exclude(Index_2, Index_1, pv{i_t}, pv{i_p}, {S_1})$ \par
$Dis{{t'}_1}, Dis{{t'}_2}, inde{x'}_1,inde{x'}_2  = chamfer\_distance({S'_1},{S'_2})$ \par
${d_{{\rm{CD}}^2}}({S_1},{S_2}) =  Mean(Dis{{t'}_1})+Mean(Dis{{t'}_2})$ 
\end{algorithm}

\subsection{The Mapping-oriented \texorpdfstring{CD$^2$}{CD2}}
Besides the distance information between the vertex set $S_2$ and ground truth point set $S_1$, additional information of mapping relation can be utilized to orchestrate the deformation process. In this subsection, we present another fine-grained CD$^2$ based on the point-vertex mapping information. Initially, we define a new data structure ${\mathbb{P}_{{V_i}}}$ to denote the set of points that consider ${V_i} \in S_2$ as their nearest vertex, i.e.,
$\psi ({P_l}) = V_i, \forall {P_l} \in {\mathbb{P}_{{{{V}}_i}}}$. A large $|{\mathbb{P}_{{V_i}}}|$ indicates that vertex ${V_i}$ is much closer to the target object than other vertices, and this vertex ${V_i}$ might not be appropriate for further deformation. From the definition of ${\mathbb{P}_{{V_i}}}$, we can find that this mapping relation contains more information and somehow includes the distance information. This benefit improves the mesh reconstruction performance, which will be extensively studied in Section \ref{sec:experiment}. Moreover, the vertex-point mapping information $|{\mathbb{P}_{{V_i}}}|$ can also be applied to measure the quality of mesh as shown in Table~\ref{tab:Vpj_quantity}.

\begin{figure}[!t]
    \centering
    \resizebox{\linewidth}{!}{
    \includegraphics[scale=0.16,trim=10 10 10 10,clip]{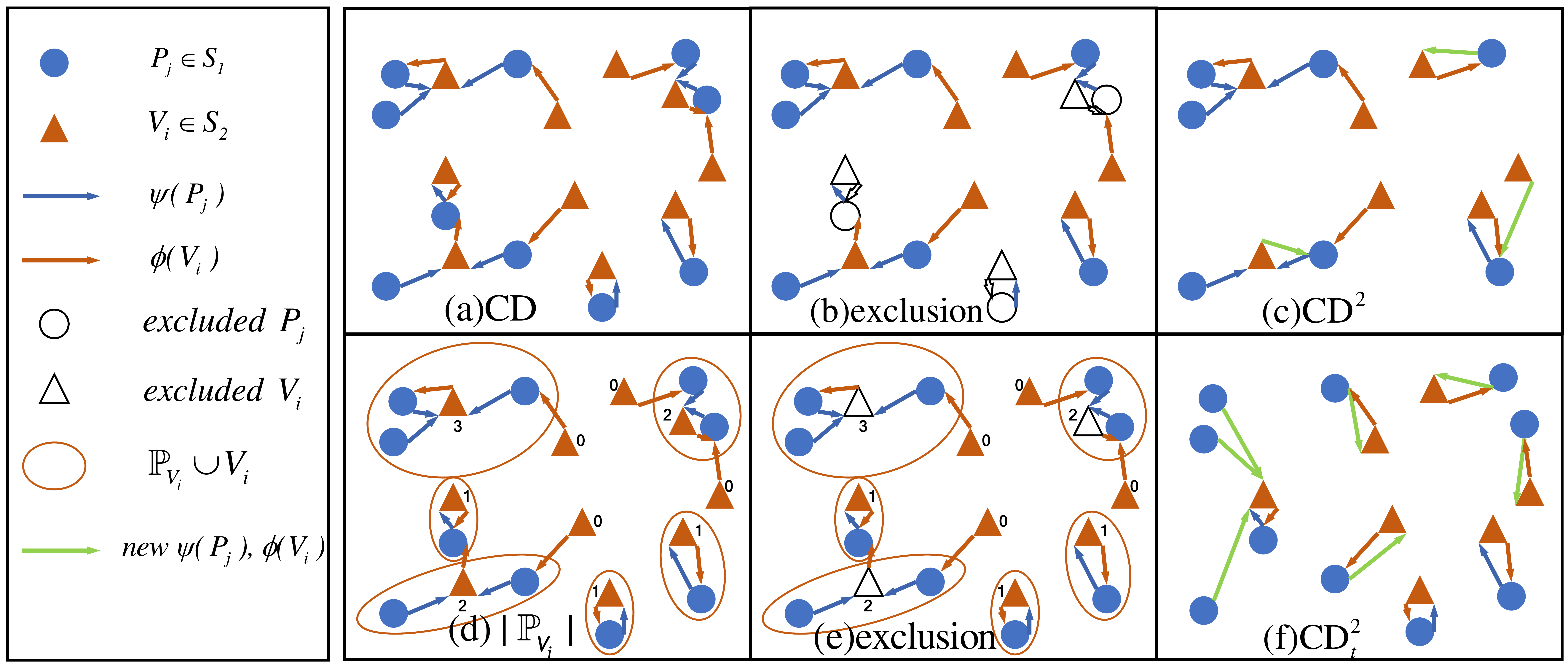}}
    \caption{The deformation demonstrations of CD, the distance-oriented ${\rm{CD}}^2$, and the mapping-oriented ${\rm{CD}}_t^2$. (a) The nearest neighbour relation in CD; (b) vertices exclusion in ${\rm{CD}}^2$; (c) the new vertex-point mapping relation of ${\rm{CD}}^2$ (${p_d}$=0.3); (d) $|{\mathbb{P}_{{V_i}}}|$; (e) vertices exclusion in ${\rm{CD}}_t^2$; and (f) the new vertex-point mapping relation of ${\rm{CD}}_t^2$ ($pv{i_t}$=1.5). We also show the legend in the left column.}
    \label{fig:Alg_explain}
\end{figure}

In the following, we use $|{\mathbb{P}_{{V_i}}}|$ to exclude some vertices and guide the model training. In this paper, we provide two specific exclusion methods with vertex-point mapping relation information. One approach ${\rm{CD}}_t^2$ eliminates vertices whose $|{\mathbb{P}_{{V_i}}}|$ is larger than a predefined threshold $pv{i_t}$, while another approach ${\rm{CD}}_p^2$ just eliminates a certain percentage $pv{i_p}$ of vertices with largest $|{\mathbb{P}_{{V_i}}}|$ value. More exclusion methods can be proposed with data structure $|{\mathbb{P}_{{V_i}}}|$, we leave them to the future work. Then, we delete points according to their $|{\mathbb{V}_{{P_j}}}|$ in $S_1$ and move residual vertices to the ground truth points in the second CD calculation. The details of our mapping-oriented methods are presented in Algorithm~\ref{algCD2_2}. Finally, we also show the deformation process of the traditional CD, the distance-oriented CD$^2$ (${p_d}$=0.3), and our mapping-oriented ${\rm{CD}}_t^2$ ($pv{i_t}$=1.5) in Fig. \ref{fig:Alg_explain}.

\section{Experiments And Evaluations}
\label{sec:experiment}
In this section, we perform extensive experiments to demonstrate the improvement of our proposals. We compare our distance-oriented CD$^2$, ${\rm{CD}}_t^2$, and ${\rm{CD}}_p^2$ with three representative baselines Atlasnet \cite{groueix_papier-mache_2018} on ShapeNet dataset \cite{chang2015shapenet}, 3DAF \cite{Wen_2022_CVPR} on ShapeNet, and Total3D \cite{nie_total3dunderstanding:_2020} on Pix3D dataset \cite{sun_pix3d:_2018}. All these works deform a template directly with CD loss. Besides these CD-related works, we also compare two recent Singed Distance Fields (SDF) works Im3d \cite{zhang2021holistic} and Tars3d \cite{duggal2022topologically} in terms of the mesh quality. Our comparison results are shown by visual demonstration and quantitative analyses with a variety of metrics including our metrics ${N_{VC}}$, ${N_{VC'}}$, ${F_{IT}}$, ${V_{IT}}$ and DPVI.

For high readability, we first summarize the definitions of these metrics. This set of metrics can evaluate mesh quality in a fine-grained and comprehensive manner.
\begin{itemize}
\item ${N_{VC}}$: the size of first-type VC vertex set. Each first-type VC vertex $V_i \in S_{V\!C}$ has the nearest neighbor $V_j \in S_{V\!C}$ and this nearest neighbor distance is smaller than the threshold ${\sigma_{V\!C}}$, i.e., $d({V_i},{V_j}) < {\sigma _{VC}}$.
\item ${N_{VC'}}$: the size of second-type VC vertex set. A second-type VC vertex $V_i \in S_{VC'} \ (S_{VC'} \subset S_{VC}$) has identical nearest ground truth points with other $V_j \in S_{VC'}$, i.e., $(d({V_i},{V_j}) < {\sigma _{VC}}) \& (\phi ({V_i}) = \phi ({V_j}))$.
\item ${F_{IT}}$: the size of IT face set $S_{IT}$ constructed by algorithm in \cite{yi2008fast}. In other words, $F_{IT} = |S_{IT}|$.
\item ${V_{IT}}$: the number of vertices which belong to the IT faces in $S_{IT}$.
\item DPVI: the number of vertices who has the same value of $|{\mathbb{P}_{{V_i}}}|$.
\item CD: a pair-wise nearest distance between two point sets. It can be computed with Eq. (1).
\item EMD: another distance metric between two point sets. It denotes the least expensive one-to-one transportation flow between two sets, which can be obtained by the optimization algorithm in \cite{wu2021density}.
\end{itemize}

\begin{table}[!t]
\caption{Statistical results of DPVI for Atlasnet, Imd3d, and Tars3d. Each value represents the average number of vertices, which has the corresponding $|{\mathbb{P}_{{V_i}}}|$, in a mesh. A large value is preferred for $|{\mathbb{P}_{{V_i}}}| = 1$, while we prefer values close to OPTA for $|{\mathbb{P}_{{V_i}}}| = 2$. Otherwise, the smaller is better.}
\label{tab:Vpj_quantity}
\centering
\resizebox{13.5cm}{!}{
\begin{tabular}{llllllllll}
\toprule
$|{\mathbb{P}_{{V_i}}}|$ & 0                & 1                & 2                & 3-10             & 11-20         & 21-30        & 31-40        & 41-50       & 51-max      \\
\midrule
$\rm{A{t_{CD}}}$ & 1154.0 	& 765.2 	& 383.3 	& 258.1 	& 1.369 	& 0.104 	& 0.019 	& 0.006 	& 0.005 
       \\ 
CD$^2$    & 1186.4 	& 740.6 	& 368.8 	& 264.0 	& 2.114 	& 0.158 	& 0.034 	& 0.009 	& 0.005 
      \\ 
${\rm{CD}}_p^2$    & 1138.5 	& \textbf{774.1} 	& 392.5 	&\textbf{ 255.8} 	& 0.984 	& \textbf{0.065 }	& \textbf{0.012} 	& 0.004 	& \textbf{0.003} 
 \\ 
${\rm{CD}}_{t}^2$    & \textbf{1137.8} 	& 773.9 	& \textbf{393} 	& 256.3 	&\textbf{ 0.929} 	& 0.066 	& 0.0124 	& \textbf{0.003} 	& \textbf{0.003} 
 \\
OPTA  & 1122.8 	& 780.8 	& 397.3 	& 261 	& 0.025 	& 0 	& 0 	& 0	& 0.000 
          \\ 
Im3d & 1329.5 	 & 601.1 	 & 327.1 	 & 299.7 	 & 3.62 	 & 0.67 	 & 0.202 	 & 0.084 	 & 0.08 

 \\
Tars3d & 1648.6 	& 329.7 	& 222.7 	& 348.5 	& 11.4 	& 0.876 	& 0.18	& 0.062 	& 0.075 
 \\

\bottomrule
\end{tabular}}

\end{table}

\subsection{Comparisons with Atlasnet on ShapeNet}
\label{sec:exper-atlas}
We first compare the performance of the distance-oriented CD$^2$, ${\rm{CD}}_p^2$, and ${\rm{CD}}_{t}^2$ with the famous scheme Atlasnet \cite{groueix_papier-mache_2018} on ShapeNet dataset \cite{chang2015shapenet}. ShapeNet dataset contains 35K manually-created 3D CAD TOMM-2022-0307, each of which involves a polygon mesh and several rendered pictures of different views. In Atlasnet \cite{groueix_papier-mache_2018}, 13 categories of ShapeNet include airplane, bench, cabinet, car, etc. Atlasnet is an impressive 3D mesh reconstruction method which is composed of 3 modules, i.e., the autoencoder trained decoder, the intermediate ResNet encoder, and the final single-view reconstruction module. Since we find that the first module of Atlasnet generates meshes with appropriate topology structures better than the other two modules, we only use the output of the first module as our baseline in this paper. In addition, multiple patches in Atlasnet are prone to generate VC and IT problems, so we use 1 patch of sphere for Atlasnet in our experiments.

We use the pretrained model provided by the Atlasnet \cite{groueix_papier-mache_2018} as a baseline $\rm{A{t_{CD}}}$, and implement our methods in Atlasnet with the following settings. For the distance-oriented method CD$^2$, we set ${p_{d}}=0.3$, ${d_T} = {10^{-7}}$. For our ${\rm{CD}}_{t}^2$, we exclude both the vertices and points with $pv{i_t} = 4$. For the $CD_{p}^2$, we set $pv{i_p} = 0.08$ for the vertex set $S_2$ and exclude $S_1$ at the ratio of 0.01. All the learning rates are ${10^{-4}}$.

\begin{figure}[!t]
    \centering
    \resizebox{\linewidth}{!}{
        \includegraphics[scale=0.32]{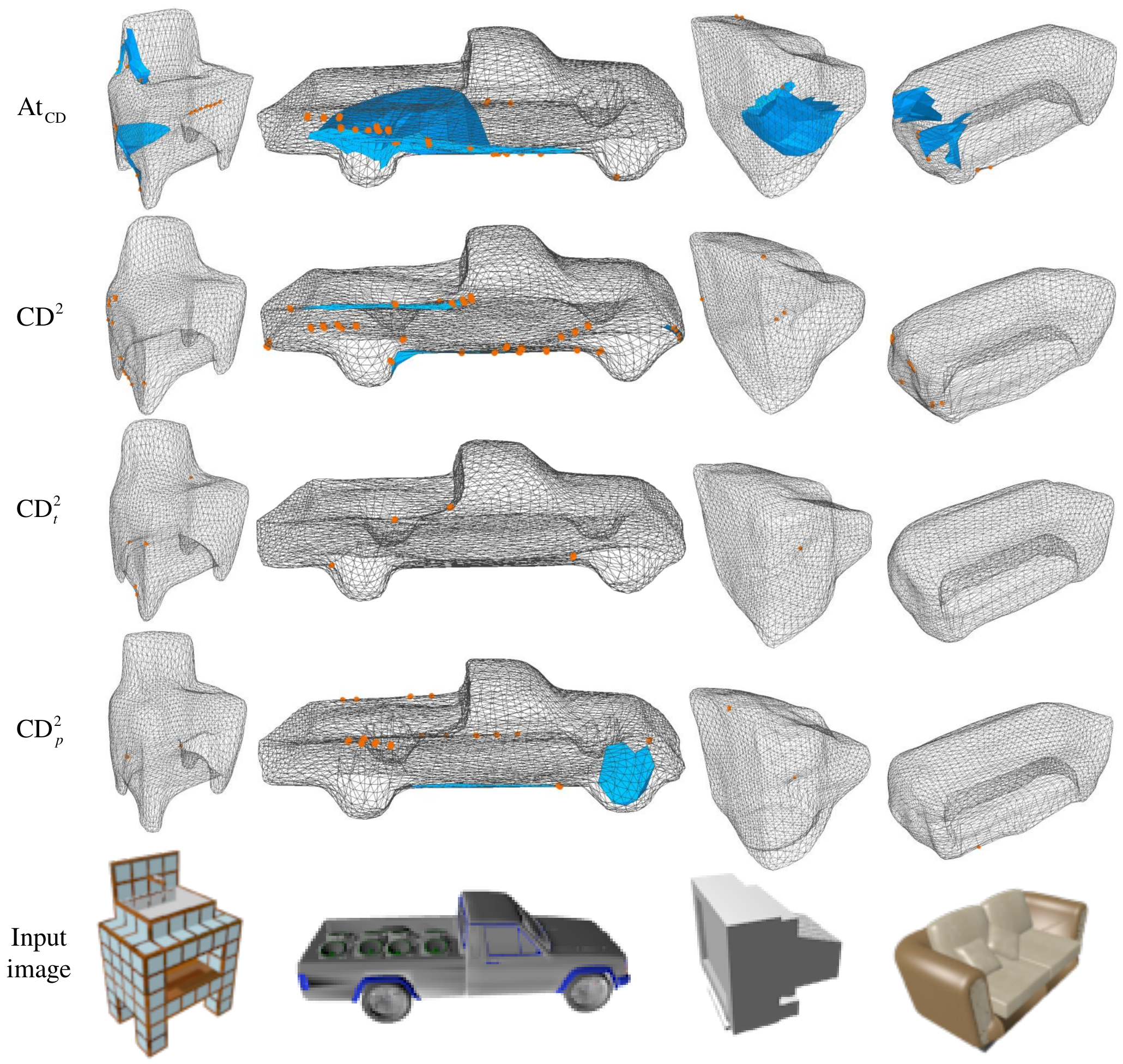}}
    \caption{The synthetic meshes on ShapeNet. We show the outputs of $\rm{A{t_{CD}}}$, the distance-oriented CD$^2$, ${\rm{CD}}_{t}^2$, and ${\rm{CD}}_p^2$ from the top row to the bottom one. IT faces are shown in blue color, while VC vertices are presented in red color.}
    \label{fig:atlas_it_VC}
\end{figure}

We use Meshlab \cite{cignoni2008meshlab} to present our visual results in Fig. \ref{fig:atlas_it_VC}. From these results, we can easily find that our proposed schemes, especially the mapping-oriented ${\rm{CD}}_p^2$ and ${\rm{CD}}_{t}^2$, generate better meshes than the baseline $\rm{A{t_{CD}}}$. For simple objects such as cabinet and sofa, our ${\rm{CD}}_{t}^2$ almost eliminates IT and VC problems. For more complicated objects like car, VC and IT problems are alleviated by our schemes, while they are quite severe in the baseline $\rm{A{t_{CD}}}$. 

The quantitative results of CD and EMD metrics are shown in Table~\ref{tab:Numeral_atlas}. Note that the CD metric is amplified by ${10^4}$ for easy comparisons in this table. The comparison results for VC and IT problems are shown in Table~\ref{tab:VCandIT}. To further demonstrate the performance improvement for VC and IT problems, we present the statistical information of the vertex-point mapping relation and define DPVI as the number of vertices who has the same value of $|{\mathbb{P}_{{V_i}}}|$. In other words, DPVI shows the information that how many ground truth points are described by a single mesh vertex. Obviously, large $|{\mathbb{P}_{{V_i}}}|$ is not appreciated by mesh reconstruction tasks. The statistical results for DPVI are presented in Table~\ref{tab:Vpj_quantity}. Our results except Table~\ref{tab:VCandIT} are computed over the entire test set. In Table~\ref{tab:VCandIT}, we average 32 TOMM-2022-0307 for VC problem and 10 TOMM-2022-0307 for IT problem due to the experimental duration consideration. We find that these results can be generalizable to the entire test set. We also present the Optimum Performance Theoretically Attainable (OPTA) for CD, EMD, and DPVI by randomly selecting 2562 points from 30K ground truth points and computing these metrics. Finally, we bold all the best results in these tables.

Our proposed schemes ${\rm{CD}}_p^2$ and ${\rm{CD}}_{t}^2$ outperform others for all the metrics except CD value. Compared with $\rm{A{t_{CD}}}$, our ${\rm{CD}}_{t}^2$ lowers EMD value by 37.9\% on average. In addition, VC vertices and IT faces are separately reduced by 31.4\% and 24.0\%. Our results also demonstrate that point-vertex mapping relation is more suitable for 3D mesh reconstruction than the distance information, which is echoed by the recent study \cite{wu2021density}. Surprisingly, our ${\rm{CD}}_p^2$ and ${\rm{CD}}_{t}^2$ output some meshes whose EMD values closely approach to the OPTA value in Table~\ref{tab:Numeral_atlas}. This indicates that results generated by ${\rm{CD}}_p^2$ and ${\rm{CD}}_{t}^2$ are almost optimal in terms of EMD metrics for categories of cabinet, car, louder speaker, sofa, telephone, etc.

\begin{table}[!t]
\caption{The quantitative results of CD and EMD on ShapeNet. From top to bottom, ${\rm{A}}{{\rm{t}}_{{\rm{CD}}}}$ represents the pretrained model offered by Atlasnet, the distance-oriented CD$^2$, ${\rm{CD}}_t^2$ and ${\rm{CD}}_p^2$ are our proposals, OPTA means the optimum performance theoretically attainable, Tars3d represents the original model provided by \cite{duggal2022topologically}. Smaller value is preferred.}
\label{tab:Numeral_atlas}
\resizebox{\linewidth}{!}{
\begin{tabular}{llllllllllllllll}
\toprule
   \rotatebox{90}{Methods} &\rotatebox{90}{Metric}  &\rotatebox{90}{airplane} & \rotatebox{90}{bench}   & \rotatebox{90}{cabinet}  & \rotatebox{90}{car}     & \rotatebox{90}{chair}          & \rotatebox{90}{display}        & \rotatebox{90}{lamp}           & \rotatebox{90}{speaker}           & \rotatebox{90}{rifle}          & \rotatebox{90}{sofa}           & \rotatebox{90}{table}          &\rotatebox{90}{telephone}            & \rotatebox{90}{vessel}         & \rotatebox{90}{all}          \\ 
    \midrule
     \multirow{2}{*}{$\rm{A{t_{CD}}}$} & CD  & \textbf{8.1}  & \textbf{10.87} & \textbf{13.45} & 14.7          & \textbf{14.4} & \textbf{13.5} & \textbf{21.9} & \textbf{20.15} & 6.02           & 13.6 & \textbf{13.9} & 10.4          & \textbf{12.2} & \textbf{13.3} \\
                       & EMD & 52.1          & 46.3          & 46.9          & 52.1          & 59.2          & 57.2          & 166.0         & 56.1          & 63.9          & 39.5          & 67.4          & 53.9          & 61.9          & 63.3          \\
\multirow{2}{*}{CD$^2$}   & CD  & 9.6           & 12.4          & 14.1          & 16.5          & 16.2          & 14.6          & 25.6          & 21.5          & 7.1           & 14.4          & 15.9          & 10.8          & 14.2          & 14.8          \\
                       & EMD & 108.6         & 176.9         & 165.2         & 93.1          & 108.6         & 164.3         & 189.8         & 178.7         & 79.8          & 244.8         & 158.7         & 75.2          & 116.3         & 143.1         \\
\multirow{2}{*}{${\rm{CD}}_{t}^2$}   & CD  & 8.6           & 11.3          & 13.5          & 14.7          & 14.9          & 14.0          & 23.8          & 20.5          & 6.3           & \textbf{13.46}          & 14.6          & 10.4          & 13.1          & 13.8          \\
                       & EMD & \textbf{33.8} & \textbf{32.5} & 31.1          & \textbf{29.2} & 40.3          & 34.4          & \textbf{82.0} & 38.3          & 50.9          & 30.2          & 41.4          & 27.7          & \textbf{38.4} & 39.3          \\
\multirow{2}{*}{${\rm{CD}}_p^2$}   & CD  & 8.4           & 10.9          & 13.46          & \textbf{14.5} & 14.6          & 13.8          & 22.7          & 20.2          & \textbf{6.0}  & 13.5          & 14.1          & \textbf{10.3} & 12.7          & 13.5          \\
                       & EMD & 41.0          & 33.2          & \textbf{30.2} & 31.0          & \textbf{37.4} & \textbf{32.7} & 83.7          & \textbf{36.6} & \textbf{46.3} & \textbf{29.6} & \textbf{38.5} & \textbf{26.9} & 38.5          & \textbf{38.9} \\
\multirow{2}{*}{OPTA}  & CD  & 4.3           & 6.7           & 11.6          & 11.1          & 9.2           & 9.5           & 5.0           & 13.6          & 2.3           & 10.6          & 8.9           & 9.2           & 5.4           & 8.3           \\
                       & EMD & 17.1          & 21.3          & 32.1          & 26.8          & 25.6          & 24.3          & 20.1          & 34.0          & 11.0          & 26.0          & 26.1          & 23.3          & 15.9          & 23.4    \\
Tars3d &EMD & 179.5 & - & - & 108.6 & 169.8 & - & - & - & - & - & - & - & - & - \\

\bottomrule
\end{tabular}
}

\end{table}

\begin{table}[!t]
\caption{The quantitative results of VC and IT problems in Atlasnet. ${N_{VC}}$ and ${N_{VC'}}$ are the average number of VC-related vertices, while ${F_{IT}}$ and ${V_{IT}}$ are the number of IT-related faces and vertices. Smaller value is preferred.}
\label{tab:VCandIT}
\resizebox{\linewidth}{!}{
\begin{tabular}{llllllllllllllll}
\toprule
 \rotatebox{90}{Method} &\rotatebox{90}{Metric}  &\rotatebox{90}{airplane} & \rotatebox{90}{bench}   & \rotatebox{90}{cabinet}  & \rotatebox{90}{car}     & \rotatebox{90}{chair}          & \rotatebox{90}{display}        & \rotatebox{90}{lamp}           & \rotatebox{90}{speaker}           & \rotatebox{90}{rifle}          & \rotatebox{90}{sofa}           & \rotatebox{90}{table}          &\rotatebox{90}{telephone}            & \rotatebox{90}{vessel}         & \rotatebox{90}{Avg.\%}          \\ 
\midrule
\multirow{4}{*}{$\rm{A{t_{CD}}}$} & ${N_{VC}}$    & \textit{188.9} & \textit{167.7} & \textit{30.5}    & \textit{30.2} & \textit{80.9}  & \textit{43.6}    & \textit{257.2} & \textit{38.9} & \textit{367.2} & \textit{35.8} & \textit{130.6} & \textit{35.5}      & \textit{142.6}  & 4.7             \\ 
                                & ${N_{VC'}}$    & 149.9          & 135.9          & 23.7             & 24.1          & 64.6           & 35.4             & 239.4          & 33.6          & 314.3          & 28.6          & 106.3          & 25.4               & 118.9           & 3.9             \\ 
                                & ${F_{IT}}$    & 321.0          & 337.3          & 74.2             & 142.9         & 343.5          & 213.6            & 229.9          & 2.7           & \textbf{486.7} & 110.1         & 809.5          & 60.0               & 172.2           & 5.0             \\ 
                                & ${V_{IT}}$    & 313.1          & 320.1          & 81.2             & 138.2         & 310.1          & 189.7            & 232.5          & 4.0           & \textbf{450.5} & 113.5         & 677.8          & 77.0               & 163.8           & 9.2             \\ 
\multirow{4}{*}{CD$^2$}    & ${N_{VC}}$    & 357.1          & 329.5          & 72.9             & 44.7          & 117.5          & 162.5            & 242.9          & 57.9          & 383.9          & 115.0         & 235.5          & 121.9              & 166.8           & 7.2             \\ 
                                & ${N_{VC'}}$    & 299.7          & 290.2          & 58.7             & 36.5          & 96.5           & 138.7            & 220.0          & 50.3          & 342.3          & 91.8          & 200.5          & 92.7               & 141.2           & 6.2             \\ 
                                & ${F_{IT}}$    & 327.7          & 424.1          & 96.5             & 46.7          & 347.7          & 176.2            & 169.7          & 15.3          & 631.9          & 73.5          & 530.2          & 59.0               & \textbf{55.5}   & 4.4             \\ 
                                & ${V_{IT}}$    & 336.6          & 407.6          & 100.7            & 46.5          & 311.4          & 184.7            & 188.4          & 20.0          & 600.8          & 77.9          & 502.5          & 69.1               & \textbf{69.7}   & 8.8             \\ 
\multirow{4}{*}{${\rm{CD}}_{t}^2$}             & ${N_{VC}}$    & \textbf{182.3} & \textbf{156.5} & \textbf{20.1}    & \textbf{7.9}  & \textbf{67.4}  & \textbf{42.5}    & \textbf{120.1} & \textbf{21.1} & \textbf{213.1} & \textbf{27.4} & \textbf{114.6} & \textbf{29.1}      & 83.1            & \textbf{3.3}    \\ 
                                & ${N_{VC'}}$    & \textbf{145.3} & \textbf{126.9} & \textbf{15.2}    & \textbf{6.1}  & \textbf{55.8}  & \textbf{33.5}    & \textbf{102.6} & \textbf{17.6} & \textbf{178.9} & \textbf{20.8} & \textbf{92.5}  & \textbf{20.8}      & 66.8            & \textbf{2.6}    \\ 
                                & ${F_{IT}}$    & \textbf{233.1} & \textbf{265.6} & \textbf{61.9}    & \textbf{12.7} & \textbf{189.5} & \textbf{137.3}   & \textbf{167.2} & 2.6  & \textbf{759.7} & 70.3          & \textbf{462.4} & \textbf{46.4}      & 152.2           & \textbf{3.8}    \\ 
                                & ${V_{IT}}$    & \textbf{250.7} & \textbf{249.8} & \textbf{63.2}    & \textbf{13.9} & \textbf{184.7} & \textbf{137.9}   & \textbf{179.2} & 4.1  & \textbf{680.2} & 75.8          & \textbf{444.2} & \textbf{54.2}      & 148.2           & \textbf{7.5}    \\ 
\multirow{4}{*}{${\rm{CD}}_p^2$}             & ${N_{VC}}$    & 223.2          & 193.3          & 28.9             & 22.6          & 90.7           & 54.8             & 147.5          & 30.9          & 233.1          & 31.8          & 123.8          & 37.1               & \textbf{72.8}   & 3.9             \\ 
                                & ${N_{VC'}}$    & 181.2          & 158.5          & 22.4             & 17.4          & 75.1           & 42.2             & 129.5          & 25.1          & 194.6          & 25.9          & 101.1          & 28.9               & \textbf{58.1}   & 3.2             \\ 
                                & ${F_{IT}}$    & 285.8          & 300.2          & 86.4             & 26.4          & 293.0          & 233.5            & 187.9          & \textbf{2.1}  & 601.6          & \textbf{59.3} & 529.5          & 59.0               & 96.9            & 4.1             \\ 
                                & ${V_{IT}}$    & 299.8          & 296.9          & 86.3             & 32.2          & 283.2          & 209.5            & 195.7          & \textbf{3.6}  & 537.4          & \textbf{63.6} & 474.9          & 60.8               & 97.4            & 7.9            \\ 
\bottomrule
\end{tabular}}
\end{table}

It should be mentioned that the classic scheme Atlasnet $\rm{A{t_{CD}}}$ only wins in CD value. This is because $\rm{A{t_{CD}}}$ use CD loss in the model training and thus CD values is almost minimized in $\rm{A{t_{CD}}}$. Our methods were trained by CD$^2$ but measured with CD, which implies that our proposals might have large CD values. This does not indicate that $\rm{A{t_{CD}}}$ is better than our schemes. Conversely, both visual results and quantitative results of all the other metrics demonstrate that our proposed schemes outperform $\rm{A{t_{CD}}}$. On the other hand, our results also demonstrate that CD metric might be inappropriate for mesh quality evaluation, while the conventional EMD and our proposed five metrics are more suitable for quality measurement.

\begin{table}[!t]
\caption{Our quantitative results in 3DAF. CD is multiplied by 1000, and EMD is multiplied by 100.}
\label{tab:3daf}
\begin{tabular}{lllllll}
\toprule
  Methods/Metrics   & CD            & EMD           & ${N_{VC}}$           & ${N_{VC'}}$         & ${V'}$           & ${F'}$           \\
 \midrule
3DAF & \textbf{4.06} & 5.56          & 141.9          & 71.9          & 1662.9          & 2951            \\
CD$^2$  & 4.71          & 5.99          & 338.2          & 181           & 1606.3          & 2777.3          \\
${\rm{CD}}_{t}^2$  & 7.1           & 5.7           & 362.2          & 160.9         & \textbf{1915.1} & \textbf{3565.6} \\
${\rm{CD}}_p^2$  & 6.06          & \textbf{5.49} & \textbf{105.5} & \textbf{48.4} & 1898            & 3532.4    \\  
\bottomrule
\end{tabular}

\end{table}

\subsection{Comparisons with 3DAF on ShapeNet}
The latest CD-based scheme 3DAF \cite{Wen_2022_CVPR}, which has drastically improved the performance of the 3D reconstruction and completion task, utilizes 3D information contained in the CAD model of ShapeNet and the global 2D feature of input image. We implement our proposal CD$^2$s in 3DAF and present our results averaged over all the categories. Both CD and EMD metrics are computed with the official codes provided by 3DAF \cite{Wen_2022_CVPR}. For the VC problem, the metrics of ${N_{VC}}$ and ${N_{VC'}}$ are calculated similarly to Section \ref{sec:exper-atlas}. Since 3DAF is proposed for point cloud reconstruction tasks, there does not exist face information in the output. We use the API \textit{create\_from\_point\_cloud\_alpha\_shape} \cite{edelsbrunner1994three} in open3d \cite{o3d_Zhou2018} to iteratively generate a mesh with the maximum face number $F_{\alpha}$. We denote the number of vertices in this mesh by ${V'}$. We employ another API \textit{create\_from\_point\_cloud\_ball\_pivoting} 
\cite{digne2014analysis,bernardini1999ball} in open3d to generate a mesh with the maximum face number $F_{bpa}$ and then obtain the average face number as ${F'}=(F_{bpa}+F_{\alpha})/2$. In short, mesh quality is evaluated by ${V'}$ and ${F'}$ in this subsection, and a large value is preferred. We use the 3DAF model trained by the author in \cite{Wen_2022_CVPR} as our baseline. Our CD$^2$s in 3DAF are trained with 100 epochs. The experimental settings are listed as follows: ${p_{d}}=0.3$ and ${d_T} = {10^{-7}}$ for the distance-oriented CD$^2$; $pv{i_t} =4$ for ${\rm{CD}}_{t}^2$; and $pv{i_p} =0.08$ for ${\rm{CD}}_{p}^2$. Other settings are identical with the original 3DAF in \cite{Wen_2022_CVPR}.

We present our results in Table \ref{tab:3daf}. Our results show that both ${\rm{CD}}_{t}^2$ and ${\rm{CD}}_p^2$ dominate 3DAF in terms of all the metrics except CD value. Our ${\rm{CD}}_p^2$ reduces VC-related vertices by 25.6\% on average. Moreover, the number of faces is increased by 19.7\% in CD$_p^2$, when we compare the proposed CD$_p^2$ with 3DAF. Our CD$_t^2$ has similar results. More faces generated by the API of open3d indicate high quality of point clouds, and thus large face number is preferred. Finally, 3DAF only wins for the CD metric mainly because CD is used as the loss function in the training.

\subsection{Comparisons with Total3D on Pix3D}
Another recent work Total3D \cite{nie_total3dunderstanding:_2020} studies 3D indoor reconstruction and scene understanding task and achieves smaller CD value on Pix3D dataset than other famous results \cite{groueix_papier-mache_2018,chen_deep_2020}. Pix3D is a large-scale dataset of well-aligned 2D images and 3D shapes. It consists of 10,069 image-shape pairs from 395 different indoor furniture in 9 categories. In this paper, we compare with two Total3D models, i.e, a pretrained model ${\rm{T3D}}_{pre}$ provided by the original Total3D \cite{nie_total3dunderstanding:_2020} and a ${\rm{T3D}_{{CD}}}$ model trained from scratch with codes offered by Total3D. We also implement the distance-oriented CD$^2$ and ${\rm{CD}}_{t}^2$ in Total3D and train them from scratch without any pretrained models. In our distance-oriented CD$^2$, we set $p_{d} = 0.6$, $d_T = 0$, and learning rate is \textbf{$10^{-4}$}. In ${\rm{CD}}_{t}^2$, $pv{i_t} = 4$, and learning rate is \textbf{$10^{-5}$}. The train-test split is similar to \cite{gkioxari_mesh_2019} for all the models, and we use the object label mapping in NYU-37 \cite{silberman2012indoor}.

\begin{figure}[!t]
\centering
\resizebox{\linewidth}{!}{
\includegraphics[scale=0.05]{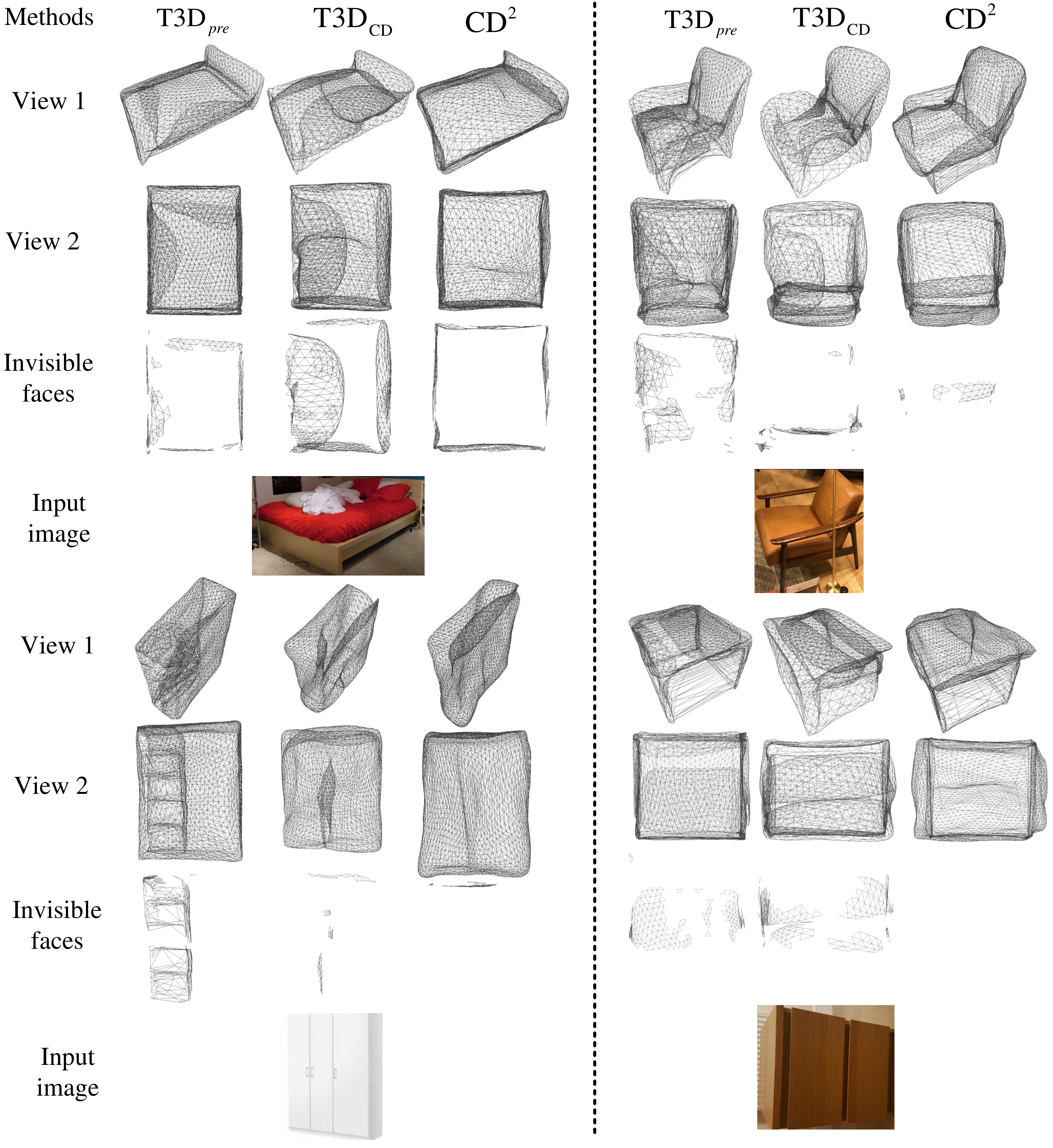}
}
\caption{Meshes in Total3D. We present the results for three methods ${\rm{T3D}}_{pre}$, ${\rm{T3D}_{{CD}}}$, and CD$^2$. For each method, we show the generated mesh from two perspectives and also provide its invisible faces.}
\label{figmesh}
\end{figure}

We first present the visual comparisons of four classes from two views in Fig. \ref{figmesh}. These results are generated by ${\rm{T3D}}_{pre}$, ${\rm{T3D}_{{CD}}}$, and the distance-oriented CD$^2$. We render these meshes with the tool of Meshlab \cite{cignoni2008meshlab}. We can easily find that meshes produced by ${\rm{T3D}}_{pre}$ and ${\rm{T3D}_{{CD}}}$ have severe IT and VC problems. For example, beds have an obvious fold on the left side; chairs have some faces folded in cushions; and both wardrobe and table have twists inside the shelf and desktop. Fortunately, our distance-oriented CD$^2$ produces meshes with much fewer twists in most classes. For instance, the mesh structure of beds generated by our CD$^2$ is also more plausible than ${\rm{T3D}}_{pre}$. Although few meshes like table are not perfect in the distance-oriented CD$^2$, they are still better than two baselines. More results are provided in Appendix \ref{apdx:t3d}. 

Another interesting result relates to invisible faces in mesh, shown in the 3rd row of each object in Fig. \ref{figmesh}. We delete all the visible faces with tools of ``select Faces in a rectangular region`` and ``select only visible`` functions in MeshLab \cite{cignoni2008meshlab}. From the results of ${\rm{T3D}}_{pre}$ and ${\rm{T3D}_{{CD}}}$, we find that there are large quantities of invisible faces in the interior mesh, and most of them have IT problems. This problem becomes even worse for the categories of bed, chair, and table. On the contrary, our distance-oriented scheme CD$^2$ has few invisible faces. We even have no invisible faces for table. These findings demonstrate our scheme generates more faces to model the surface of objects and outputs well-structured meshes.

\begin{table*}[!t]
\caption{The quantitative results of VC and IT problems in Total3D. The last column is the average percentage over all the categories. The smaller is better. Each sample consists of 2,562 vertices and 5,120 faces.}
\label{tab:vcitTotal3D}
\centering
\resizebox{\linewidth}{!}{
\begin{tabular}{llllllllllll}
\toprule
\rotatebox{90}{Method}                                  & \rotatebox{90}{Metric} & \rotatebox{90}{bed}            &\rotatebox{90}{bookcase}       & \rotatebox{90}{chair}          & \rotatebox{90}{desk}            & \rotatebox{90}{msic}           & \rotatebox{90}{sofa}           & \rotatebox{90}{table}          & \rotatebox{90}{tools}           & \rotatebox{90}{wardrobe}       & \rotatebox{90}{Avg.\%}         \\ 
\midrule
\multirow{4}{*}{${\rm{T3D}}_{pre}$}                    & $N_{VC}$    & 263.3          & 523.7          & \textbf{489.5} & \textbf{644.8}  & 440.5          & 180.3          & \textbf{347.9} & 461.5           & \textbf{190.9} & 15.36          \\  
                                        & $N_{VC'}$   & 200.0          & 446.5          & \textbf{423.2} & \textbf{559.8}  & 381.8          & 129.4          & \textbf{285.0} & 403.7           & \textbf{149.2} & 12.92          \\  
                                        & $F_{IT}$    & 893.1          & 1265.5         & 1549.5         & 2283.0          & 1334.4         & 777.6          & 1429.5         & 2834.2          & 581.9          & 28.10          \\  
                                        & $V_{IT}$    & 830.1          & 1174.7         & 1281.9         & 1763.6          & 1133.5         & 729.8          & 1211.9         & 2054.7          & 528.9          & 46.44          \\ 
\multirow{4}{*}{${\rm{CD}}^2$}  & $N_{VC}$    & 464.8          & 892.4          & 923.0          & 1051.6          & 574.9          & 228.8          & 731.6          & 500.8           & 322.8          & 24.68          \\  
                                        & $N_{VC'}$   & 389.2          & 816.2          & 832.1          & 974.8           & 520.9          & 174.5          & 630.6          & 431.5           & 259.7          & 21.81          \\  
                                        & $F_{IT}$    & 909.6          & 1195.7         & 1479.4         & 1777.6          & 682.5          & 838.0          & 1238.4         & 1858.6          & 511.8          & 22.77          \\  
                                        & $V_{IT}$    & 860.4          & 1001.3         & 1243.9         & 1484.5          & 648.2          & 809.0          & 1053.6         & 1445.2          & 497.9          & 39.22          \\ 
\multirow{4}{*}{${\rm{CD}}_{t}^2$} & $N_{VC}$    & \textbf{148.0} & \textbf{456.5} & 590.5          & 671.7           & \textbf{175.5} & \textbf{74.7}  & 526.4          & \textbf{151.8}  & 287.8          & \textbf{13.37} \\  
                                        & $N_{VC'}$   & \textbf{121.2} & \textbf{419.1} & 521.8          & 610.8           & \textbf{152.4} & \textbf{58.1}  & 459.3          & \textbf{124.7}  & 243.5          & \textbf{11.76} \\  
                                        & $F_{IT}$    & \textbf{344.4} & \textbf{642.2} & \textbf{941.9} & \textbf{1351.2} & \textbf{351.5} & \textbf{275.8} & \textbf{725.9} & \textbf{1002.0} & \textbf{392.7} & \textbf{13.08} \\  
                                        & $V_{IT}$    & \textbf{356.4} & \textbf{560.9} & \textbf{867.6} & \textbf{1171.4} & \textbf{352.3} & \textbf{285.7} & \textbf{678.7} & \textbf{946.4}  & \textbf{381.3} & \textbf{24.29} \\ 
\bottomrule
\end{tabular}
}
\end{table*}

\begin{table*}[!t]
\caption{EMD results in Total3D. All the presented results have been divided by 100, and the last column shows the average value for each method. The smaller is better.}
\label{tab:EMD_t3d}
\centering
\begin{tabular}{lllllllllll}
\toprule
\textbf{} & bed           & \textbf{bookcase} & \textbf{chair} & desk          & msic           & \textbf{sofa} & table         & tools          & wardrobe      & Avg.           \\ 
\midrule
${\rm{T3D}}_{pre}$       & \textbf{2.71} & 12.88             & \textbf{4.81}  & \textbf{5.98} & 11.69 & 2.14          & \textbf{6.52} & 13.67 & 7.42          & 7.54          \\ 
CD$^2$      & 3.64 & 11.72             & 8.00           & 7.94 & 10.84          & \textbf{1.88} & 6.55          & \textbf{8.33}  & \textbf{2.89} & 6.87          \\ 
${\rm{CD}}_{t}^2$        & 3.85          & \textbf{9.99}     & 6.07           & 6.42          & \textbf{10.30} & 2.24          & 6.89          & 8.95           & 4.86          & \textbf{6.62} \\ 
Im3d & 1.83 	& 0.91 	& 1.74 	& 2.01 	& 2.82 	& 0.82 	& 2.71 	& 0.67 	& 0.74 	& 1.58 
 \\
\bottomrule
\end{tabular}
\end{table*}

We also perform quantitative comparisons among three models, i.e., the official ${\rm{T3D}}_{pre}$, our distance-oriented CD$^2$, and ${\rm{CD}}_{t}^2$. The results of IT and VC problems in Total3D are shown in Table \ref{tab:vcitTotal3D}. Our experiment results demonstrate the proposed ${\rm{CD}}_{t}^2$ outperforms ${\rm{T3D}}_{pre}$ again, especially for the IT problem. In detail, the output mesh in ${\rm{T3D}}_{pre}$ has more than 28.10\% faces and 46.44\% vertices of IT problems. However, the IT-related faces in our ${\rm{CD}}_{t}^2$ are only half of ${\rm{T3D}}_{pre}$. In addition, our ${\rm{CD}}_{t}^2$ is better than the distance-oriented CD$^2$, which is consistent with the previous results in Atlasnet in Section \ref{sec:exper-atlas}. Finally, ${\rm{CD}}_{t}^2$ has more VC vertices for four categories, i.e., chair, desk, table, and wardrobe, compared with ${\rm{T3D}}_{pre}$. This is because we move some invisible vertices to the surface of object, which efficiently reduces the IT problem and shortens the vertex distance. But this phenomenon only exists in a few special categories, and our schemes alleviate VC and IT simultaneously for most categories.

We show the EMD results for ${\rm{T3D}}_{pre}$, our distance-oriented ${\rm{CD}}^{2}$, and ${\rm{CD}}_{t}^2$ in Table \ref{tab:EMD_t3d}. Our schemes achieve smaller EMD results than ${\rm{T3D}}_{pre}$ on average. For the classes of bed, chair, desk, and table, the Total3D baseline ${\rm{T3D}}_{pre}$ performs slightly better than our schemes. The reasons are similar to the results of VC explained above. In other words, meshes with more IT vertices may have smaller VC and EMD metrics since fewer vertices are located at the surface of meshes. There is a tradeoff between these metrics, and the study of their relationship is left for future work.

We also use quantitative results to confirm some observations from visual comparisons. We randomly select ten TOMM-2022-0307 from some categories and calculate the average number of invisible faces and show the results in Table \ref{facesunused}. From our results, we ensure that a mesh produced by the distance-oriented CD$^2$ has fewer invisible faces than ${\rm{T3D}_{{CD}}}$ and ${\rm{T3D}}_{pre}$. These results verify that more faces contribute to the surface of objects in our scheme, which benefits the mesh generation. 

\begin{figure}[!t]
\centering
\resizebox{\linewidth}{!}{
\includegraphics[scale=0.21]{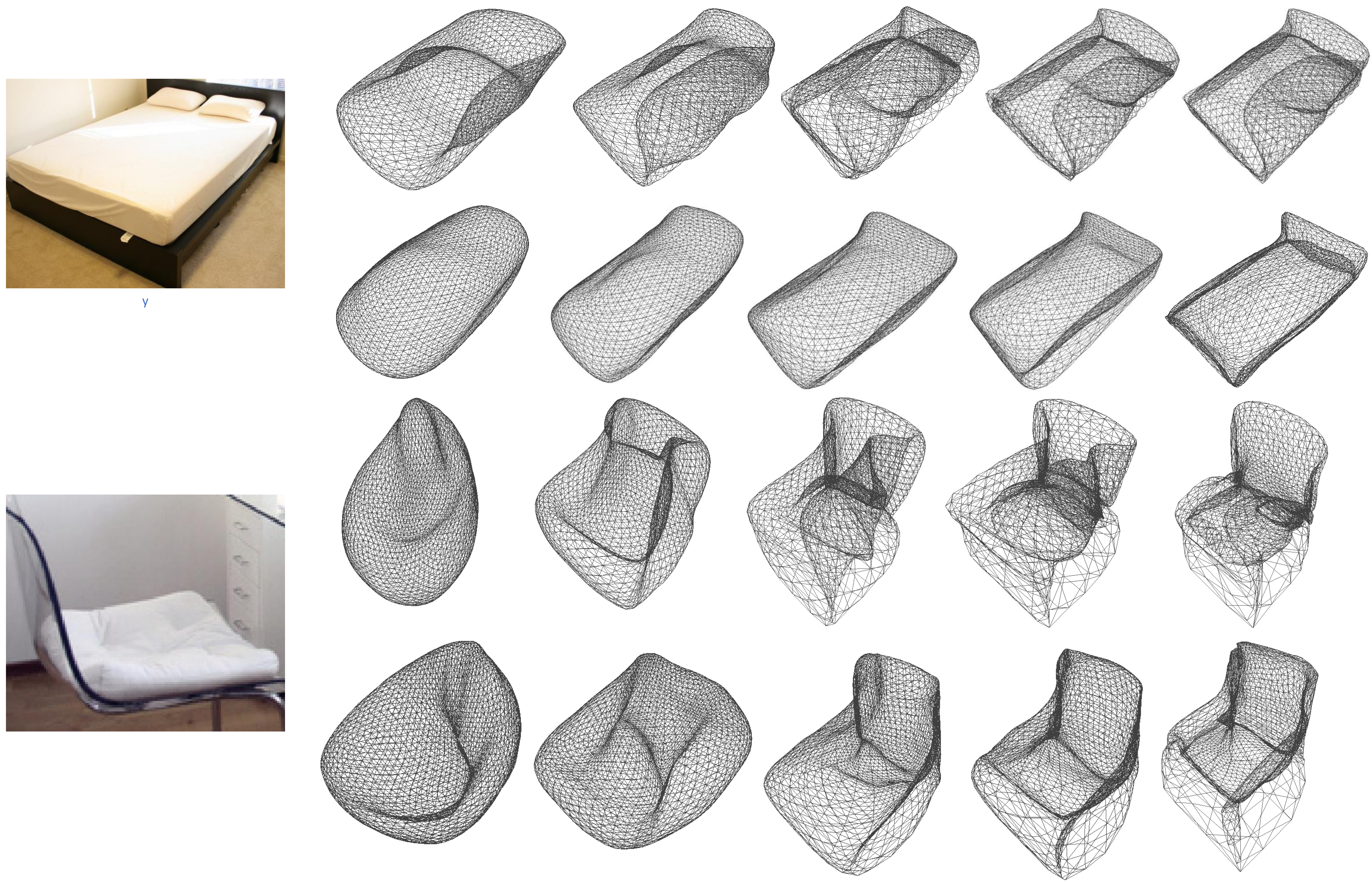}
}
\caption{Mesh deformation during learning progress. Meshes are generated by ${\rm{T3D}_{{CD}}}$ for the 1st and 3rd row, while other results are obtained from CD$^2$.}
\label{MGNprogress} 
\end{figure}

\begin{table}[!t]
\caption{The quantity of invisible faces in generated meshes. Smaller values are preferred.}
\label{facesunused}
\centering
\begin{tabular}{ccccc}
   \toprule
Method  & Bed & Bookcase & Chair & Desk\&Table \\
   \midrule
${\rm{T3D}}_{pre}$     & 604 & 890.6    & 473.8 & 377.8  \\ 
${\rm{T3D}_{{CD}}}$ & 979 & 1023.4   & 499.2 & 428.4 \\
CD$^2$     & \textbf{310} & \textbf{386.8}    & \textbf{139}   & \textbf{17.2} \\ 
 \bottomrule
\end{tabular}
\end{table}

In this subsection, we also demonstrate the deformation process of two TOMM-2022-0307 in Fig. \ref{MGNprogress}. In the 1st and 3rd rows, we can find the generation process of folds and twists caused by the CD in Total3D. However, our distance-oriented CD$^2$ in the 2nd and 4th rows succeeds in approaching the target object by performing the fine-grained deformation, instead of the brute-force nearest neighbour search procedure. Fig. \ref{MGNprogress} is a significant proof of performance improvement in our CD$^2$.

\subsection{Comparisons with SDF-based schemes}
\begin{figure}[!t]
    \subfloat[visual results with Tars3d.]{\label{fig-tars3d}\includegraphics[scale=0.45]{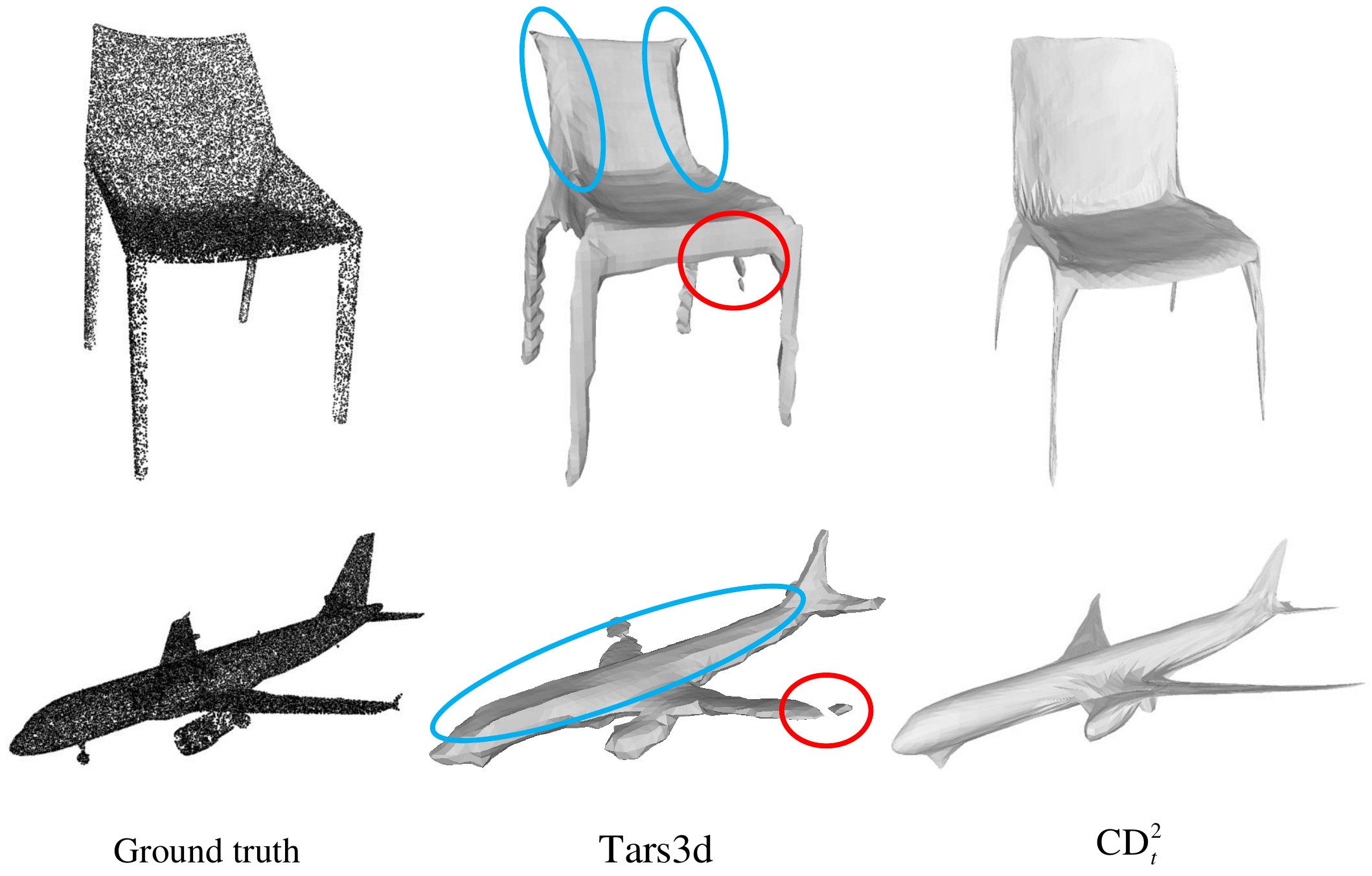}}
    \rule{0.8pt}{170pt}
    \subfloat[visual results with Im3d.]{\label{fig-im3d}\includegraphics[scale=0.45]{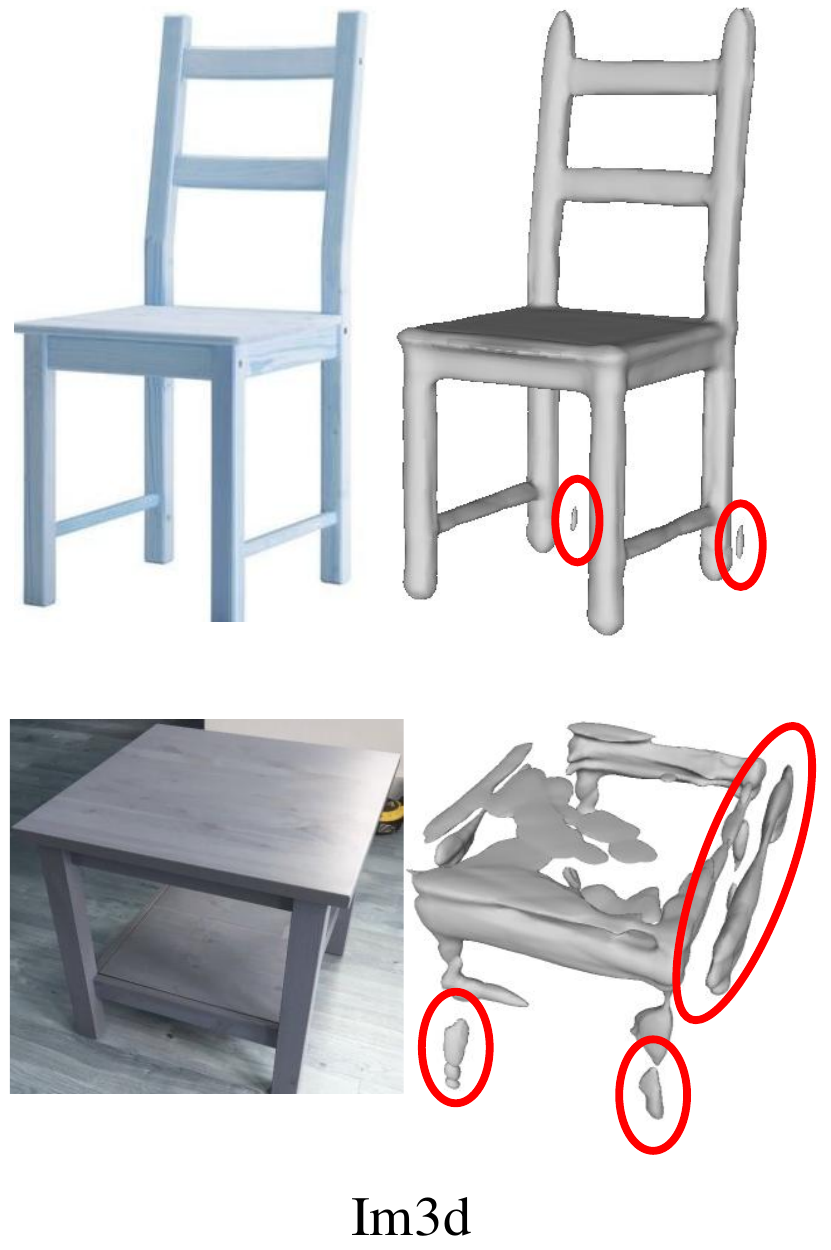}}     \caption{Mesh reconstruction results of ${\rm{CD}}_{t}^2$ with Atlasnet, Tars3d, and Im3d. We use blue circles to show inappropriate reconstructed meshes and red circles to highlight the impending parts of meshes.}
    \label{fig:tars3d_im3d}
\end{figure}

Recently, some implicit 3D representation methods like SDF-based schemes are becoming popular and attracting much attention in academia. In this paper, we compare our work with two state-of-the-art implicit representation methods Im3d \cite{zhang2021holistic} and Tars3d \cite{duggal2022topologically}. In Table \ref{tab:Numeral_atlas}, compared with $\rm{A{t_{CD}}}$, Tars3d achieves larger EMD values for all three categories. Meanwhile, our ${\rm{CD}}_{t}^2$ has much smaller EMD values than Tars3d. Note that our ${\rm{CD}}_{t}^2$ is trained on all 13 categories, while Tars3d is trained separately for each category. If our mapping-oriented scheme ${\rm{CD}}_{t}^2$ is trained for each individual category, the performance gap will be further enlarged. We leave this extension to the future work. In addition, we find that Tars3d cannot model objects accurately for some specific parts in Fig. \ref{fig:tars3d_im3d}. For instance, meshes of airplane cabin and chair backrest have many defects shown in blue circles, and thus they are not properly reconstructed. However, Im3d performs better than Total3D and our CD$^2$s in terms of EMD metrics, as shown in Table \ref{tab:EMD_t3d}. The visual performance improvement of Im3d can also be found in Fig. \ref{fig:tars3d_im3d}(b), and the mesh of chair seems to be perfect among these three comparable schemes.

We also show the DPVI results of these two SDF-based schemes in Table \ref{tab:Vpj_quantity}, and find that they both perform worse than our ${\rm{CD}}_{t}^2$ and ${\rm{CD}}_p^2$. In practice, SDF-based schemes might introduce additional problems like some impending parts shown in red circles in Fig. \ref{fig:tars3d_im3d}. Even worse, SDF-based schemes might generate messy and implausible meshes when the input image contains dim light or cluttered background. We show this problem with a table sample in Fig. \ref{fig:tars3d_im3d}(b). These two problems contribute to the undesirable DPVI. In sum, our schemes outperform two SDF-based schemes in terms of DPVI and dominate Tars3d in terms of EMD.

\subsection{The time consumption of CD, EMD, and CD$^2$}
\label{sec:time_c}
\begin{figure}[!t]
    \centering
    \resizebox{10.5cm}{!}{
     \includegraphics[scale=0.4]{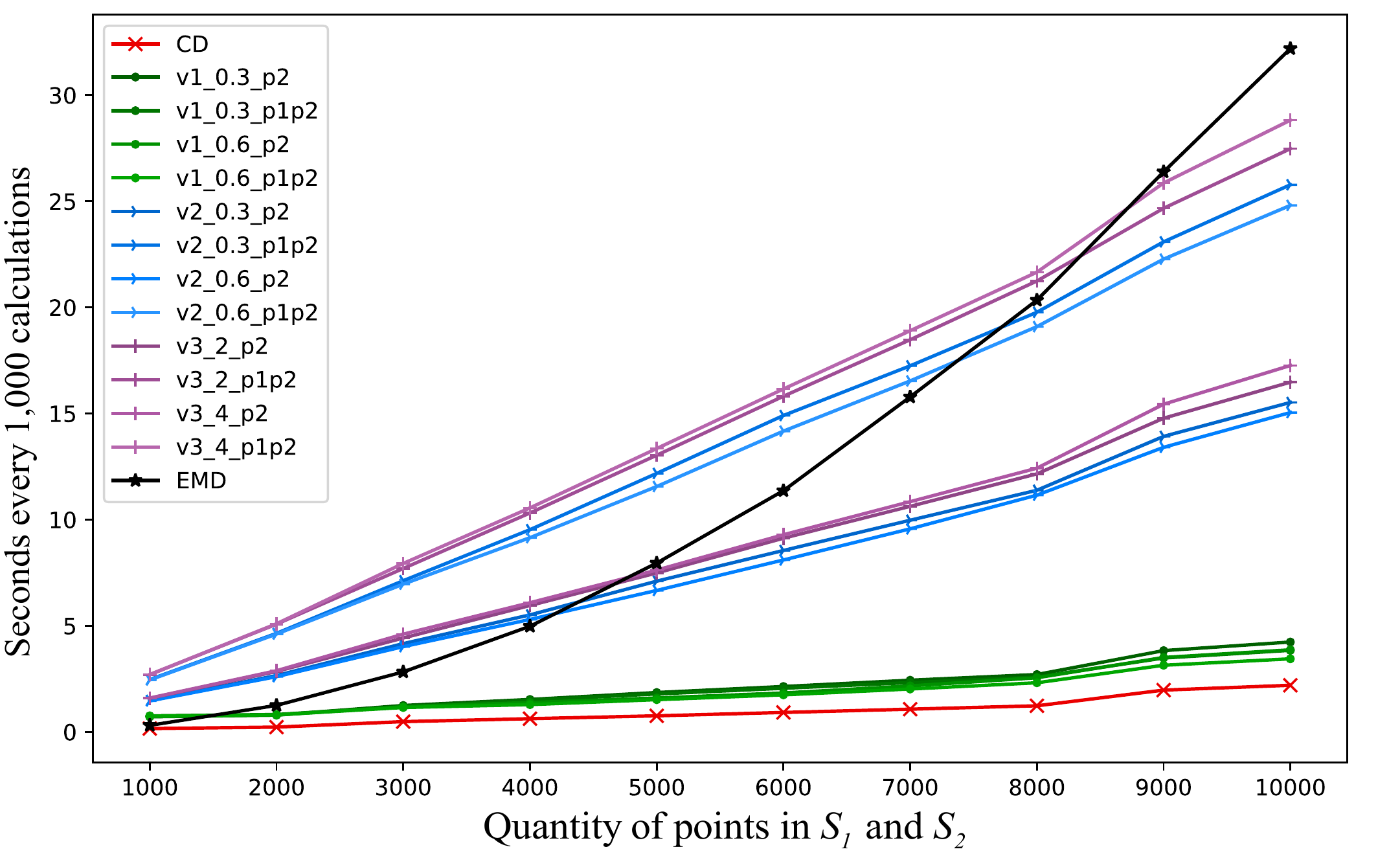}}
    \caption{The time consumption of CD, EMD, and CD$^2$s. The EMD result is divided by 45. In this legend, v1, v2, and v3 are separately the distance-oriented CD$^2$, ${\rm{CD}}_p^2$, and ${\rm{CD}}_{t}^2$. The following decimals are the exclusion percentage (${p_{d}}$ and $pv{i_p}$), and the integers indicate the exclusion threshold ($pv{i_t}$). p2 means the vertex exclusion in $S_2$, and p1p2 means the point-vertex exclusion in $S_1$ and $S_2$ simultaneously.}
    \label{fig:time_consumption}
\end{figure}

We show the time consumption of three metrics CD, EMD, and CD$^2$s in distance calculations between two point sets $S_1$ and $S_2$ for 1,000 times. The results are shown in Fig. \ref{fig:time_consumption}. Except for the EMD metric, all the results can be modeled as an approximately linear increase function. In addition, we can find that the time consumption of the distance-oriented CD$^2$ is close to that of CD, but our ${\rm{CD}}_p^2$ and ${\rm{CD}}_{t}^2$ have larger slopes because mapping-related vertex elimination operation consumes more time than distance-based method. On the other hand, the EMD metric has an exponential rise in time consumption, which is much larger than CD-based metrics. When there are 10,000 points in the point sets, the computation time is more than 1.20s for each EMD calculation. Compared with EMD, our CD$^2$s are computation-efficient. 

\section{Related Work}
\label{sec:relatedwork}
In this section, we review some representative methods and machine-learning-enabled techniques in 3D reconstruction. Then, some major metrics are surveyed for the loss function, and we mainly discuss the CD loss in the explicit mesh deformation process.

\subsection{3D Reconstruction with Machine Learning} 
There exists a variety of studies in the field of 3D reconstruction from RGB images using promising machine learning techniques. In these studies, the output format of 3D objects includes point cloud \cite{wu2021density,lin2018learning,pala_reconstructing_2019}, voxel \cite{shin_3d_2019,wang20173densinet,xie_pix2vox:_2019}, primitives \cite{groueix_papier-mache_2018}, mesh \cite{nie_total3dunderstanding:_2020}, etc. Among them, mesh appeals to complicated topology and can represent any 3D object accurately and efficiently. For the 3D mesh reconstruction, one line of studies generates mesh from SDF \cite{dai_shape_2017} and other implicit surfaces \cite{oechsle_unisurf:_2021}. For instance, Zhang et al. learned a 3D shape as local implicit information and then decoded it to SDF values \cite{zhang2021holistic}. Duggal and Pathak mapped the input 3D point clouds to canonical space and then employed DeformNet and shape generator to generate SDF \cite{duggal2022topologically}. In this category of research, the marching cube algorithm \cite{lorensen1987marching} is adopted to obtain the final mesh from the intermediate SDF. Meanwhile, another line of impressive studies explicitly and directly reconstructs a mesh from a template with deformation methods, which is the focus of this paper. Early works can only produce a category-specific mesh by transforming a learned mean shape \cite{tulsiani_learning_2017}. Recently, some studies have been proposed to yield meshes by deforming different initial templates. For example, Pontes et al. obtained a compact mesh representation by deforming a cube shape \cite{pontes2018image2mesh}. Gkioxari et al. converted the coarse voxel to mesh by refining the vertices and edges with graph convolution network \cite{gkioxari_mesh_2019}. Wang et al. deformed an ellipsoid to the target object by utilizing the perceptual features from images \cite{wang2018pixel2mesh}. In \cite{groueix_papier-mache_2018}, Groueix et al. supported mesh deformation with different amounts of spheres or squares called primitives. Besides these results, Huang et al. applied multi-view renderings with a 3D human template to a deformation model and achieved photo-consistency human meshes \cite{yu_multi-view_2022}. Chen et al. folded a 2D lattice to the target object and then learned a pairwise mapping between the 3D data \cite{chen_deep_2020}. In \cite{nie_total3dunderstanding:_2020}, Nie et al. further explored the main differences and impacts between cutting edges and removing faces in the topology modification operations \cite{pan_deep_2019}. 

\subsection{Metrics of 3D Reconstruction} 
CD \cite{barrow1977parametric}, EMD \cite{rubner2000earth}, and Intersection over Union (IoU) are widely-adopted metrics to measure the distance between reconstructed mesh and the target object point set in current research. Since IoU is not differentiable, most works use CD and EMD as their loss functions in deep learning models as well as metrics to evaluate the quality of reconstructed mesh \cite{nie_total3dunderstanding:_2020,wang2018pixel2mesh,gkioxari_mesh_2019,groueix_papier-mache_2018}. CD calculates the nearest pair-wise distance from one point set to the other one, while EMD relies on solving an optimization problem to obtain the best mapping function from one set to the other set. Although EMD is more faithful than CD, the computation cost of CD (i.e., ${\rm{O}}(N{\rm{ \times }}\log (N))$) is much smaller than that of EMD (i.e., $O({N^3}{\rm{ \times }}\log (N))$) \cite{shirdhonkar2008approximate}. In addition, CD can be calculated in parallel and further accelerated by KD-tree  \cite{bentley_multidimensional_1975} and Octree \cite{1980Octree}. Besides the above metrics, DPDist \cite{urbach2020dpdist} and the sliced Wasserstein distance \cite{bonneel2015sliced} are proposed in recent years, but they are not the mainstream metrics in various tasks. In this paper, we concentrate on the preferable CD and improve it in the model training and evaluation.

\subsection{Chamfer Distance} 
Some severe problems have been identified for the CD loss in deep learning model. For instance, Achlioptas et al. discovered that CD is prone to generate outputs with points crowded in the area with the highest occupancy probability (e.g. the seat of chairs), and there exists an imbalance problem between two summands $d_{C\!D1}({S_1},\!{S_2})$ and $d_{C\!D2}({S_1},\!{S_2})$ \cite{achlioptas2018learning}. Li et al. showed that CD suffers from the local optimum problem in the nearest neighbor search of a finger moving example \cite{li_lbs_2019}. In \cite{jin2020dr}, Jin et al. found that thickening, elongating, and shortening four legs of a chair result in a larger CD value than removing any leg, which implies that CD is not faithful visually and structurally. By comparing the loss of CD with Mean Squared Error (MSE) in model training, Wagner et al. discovered that optimizing CD directly would neglect some details, especially for densely-sampled point cloud \cite{wagner_neuralqaad:_2022}. In \cite{paschalidou_neural_2021}, Paschalidou et al. pointed out that Atlasnet with optimized CD would generate zero-volume primitives and faces with inverted normal. Their observations partially and implicitly demonstrate the self-intersection problem induced by CD. In a nutshell, the IT and VC problems caused by CD loss have not been explicitly discovered before in the deformation process.

Meanwhile, some variant CDs have been proposed to improve the deformation performance. Li et al. proposed a structured CD scheme which divides 3D objects into several regions and computes the nearest neighbour search in a region-to-region manner other than an all-to-all manner \cite{li_lbs_2019}. Lim et al. provided a sharper version of CD, which summarizes the $p$ power of each distance over all the points and then computes its $p$th root \cite{lim_convolutional_2019}. Their scheme punishes points with large errors more heavily than before. In \cite{wang2020deep}, Wang et al. proposed an adaptive CD loss for better aligning partial shapes. In detail, they obtained the overlapping subset for each point set with a pre-defined distance threshold and then applied CD to these two overlapping subsets. In the optimization process, the pre-defined distance threshold is chosen adaptively. Li et al. presented a probabilistic CD loss for accurate key point localization in \cite{li_usip:_2019}. Chen et al. proposed an augmented CD by only choosing the largest term each time \cite{chen_deep_2020}. The most relevant work is \cite{wu2021density}, where Wu et al. proposed a Density-aware CD (DCD) for balanced point cloud reconstruction. DCD considers each vertex equally in terms of its queried frequency and then lessens the impact of outliers. In comparison, our work concentrates on mesh reconstruction and treats mesh vertices unequally with moderate speeds and correct directions. In this way, we endeavor to orchestrate the deformation process to mitigate the IT and VC problems, which is another contribution of this paper.

Some additional constraints are added to the CD loss function to improve mesh quality. These auxiliary constraints include factual constraints and hypothetical assumptions. For the first category, the edge distance constraint avoids merging adjacent edges \cite{gkioxari_mesh_2019}; Laplacian loss \cite{desbrun_implicit_1999} and normal loss \cite{wang2018pixel2mesh} both impose the smoothness of mesh; boundary regularization enforces smoothness and consistency of boundary curves \cite{pan_deep_2019}; and locality loss penalizes points which locate outside of a small spatial region \cite{lim_convolutional_2019}. For the second group of auxiliary constraints, unsup3d in \cite{wu_unsupervised_2021} assumed that all human faces were symmetric and added this symmetry assumption to the loss function. Alex et al. studied how the physics-inspired prior knowledge impacts the model to learn dynamics, hypothesizing that all the objects in video datasets obey the law of gravity \cite{botev2021priors}. Michael et al. even argued that deep-learning-enabled methods cannot survive without prior knowledge \cite{oechsle_unisurf:_2021}. Finally, it should be noted that these auxiliary constraints are orthogonal to our paper.

\section{Conclusion}
\label{sec:conclusion}
In this paper, we have identified that CD loss may introduce VC and IT problems in explicit mesh reconstruction and then provided fine-grained schemes CD$^2$s to orchestrate the deformation of vertices and alleviate these two problems. Our schemes first compute CD loss to exclude aggressively-deformed vertices with distance or mapping information and then move an adaptive vertex subset to the ground truth points with a second CD calculation. We have implemented our CD$^2$s in three representative CD-based schemes, Atlasnet, 3DAF, and Total3D, on two datasets ShapeNet and Pix3D. We compare our CD$^2$s with these baselines in terms of various metrics, most of which are proposed by this paper. Extensive experimental results show that our proposals outperform baselines for almost all the categories and produce more plausible and well-structured meshes. We also compare two implicit SDF-based reconstruction schemes with our proposals and find that our CD$^2$s dominate SDF-based schemes in terms of DPVI metric. Some theoretical studies of VC and IT problems will benefit 3D mesh reconstruction, and we leave them in future work.

\newpage

\appendix 

\section{Appendix}
\subsection{More visual results in Total3D}
\label{apdx:t3d}
\begin{figure}[!h]
\centering
\resizebox{\linewidth}{!}{
\includegraphics[scale=0.06]{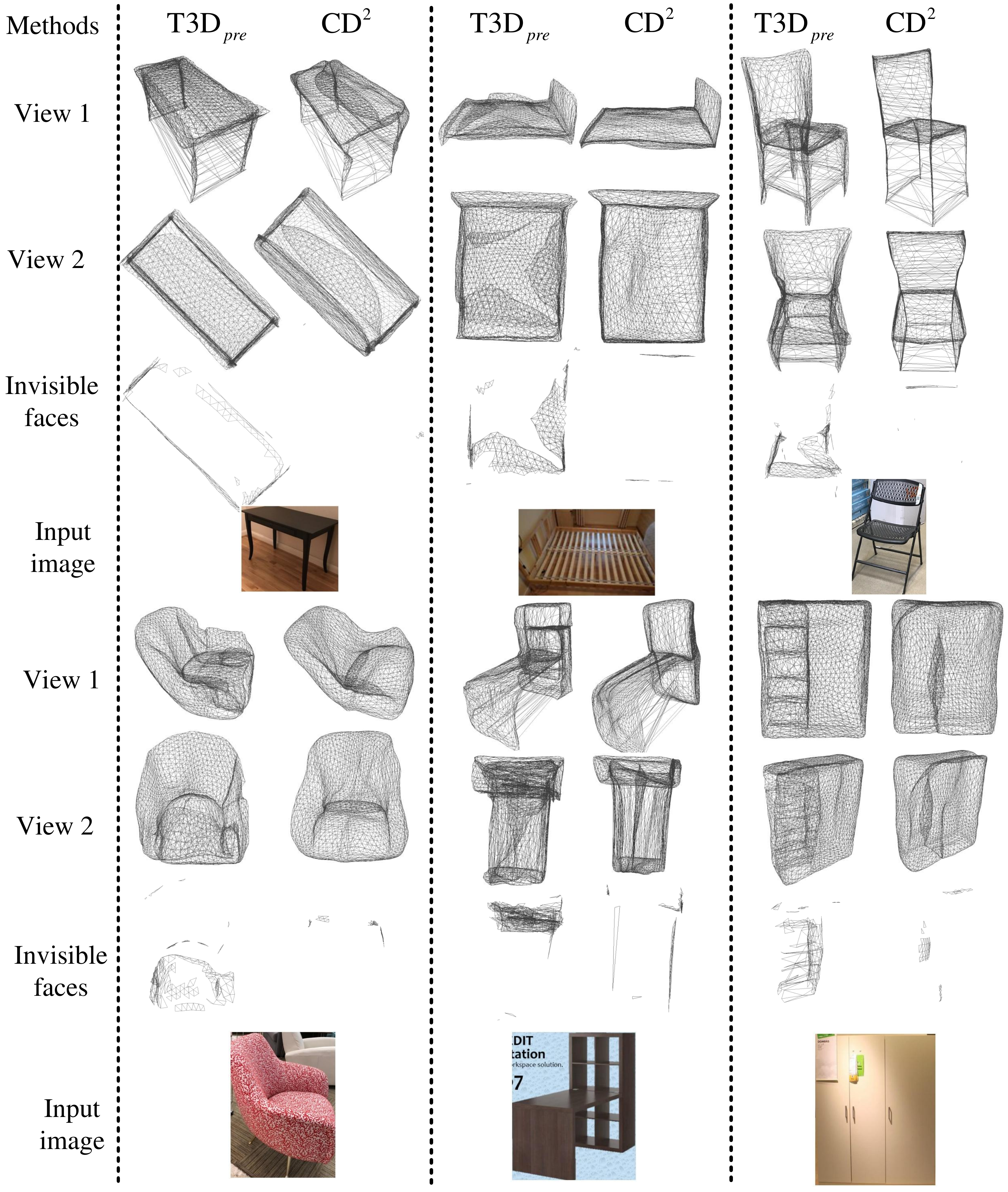}
}
\caption{More meshes in Total3D. We compare the output meshes generated by T3D$_{pre}$ and our CD$^2$ from two perspectives and invisible faces.}
\label{figmesh_t3d_appendix}
\end{figure}

In Fig. \ref{figmesh_t3d_appendix}, we show more reconstructed meshes in Total3D with the pretrained ${\rm{T3}}{{\rm{D}}_{pre}}$ and our distance-oriented CD$^2$.

\subsection{More visual results in Atlasnet}
\begin{figure}[!h]
    \centering
    \resizebox{\linewidth}{!}{
    \includegraphics[scale=0.31]{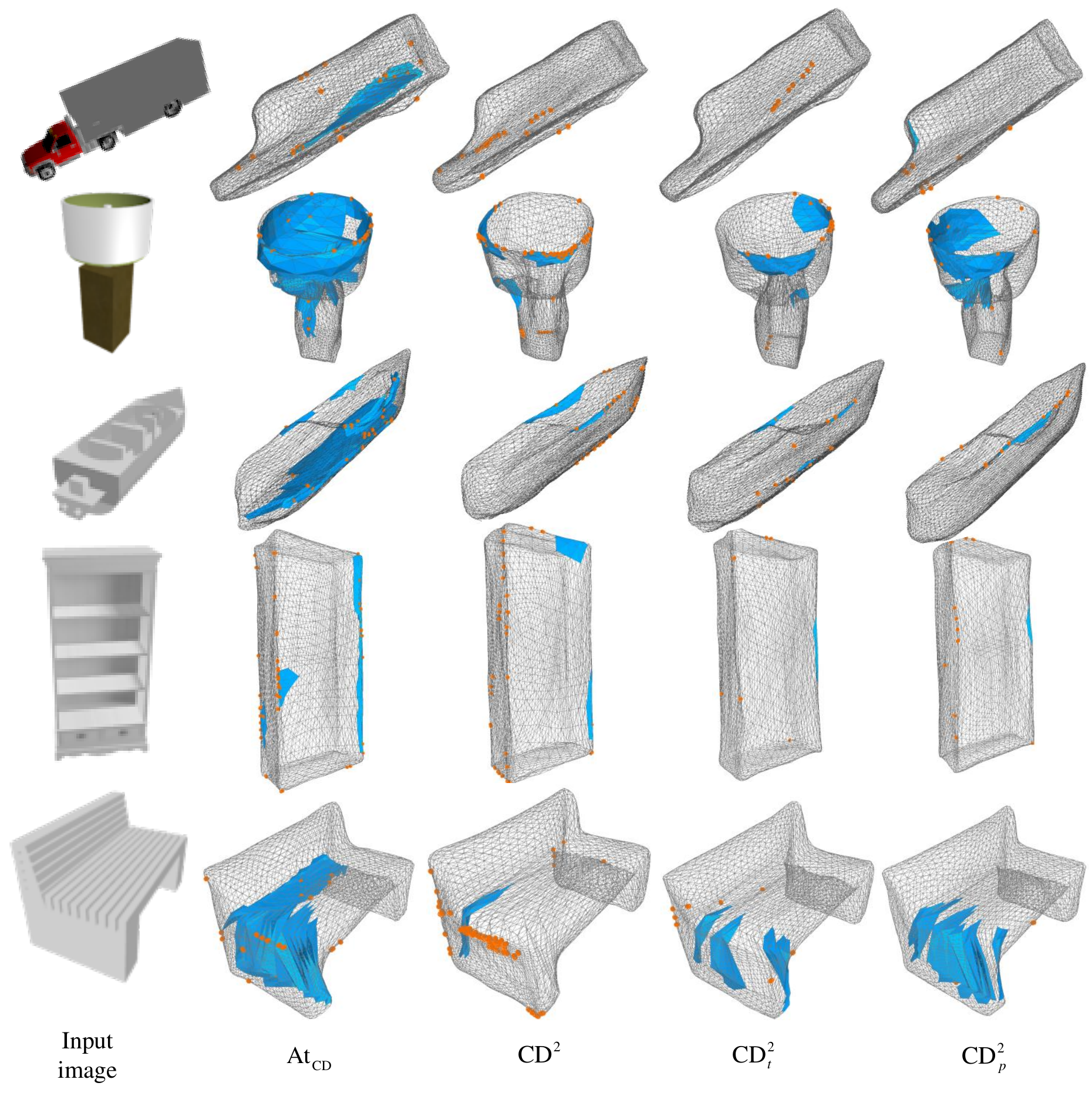}}
    \caption{More meshes in Atlasnet. IT faces are shown in blue color, while VC vertices are presented in red color.}
    \label{fig:atlas_it_VC_appendix}
\end{figure}

In Fig. \ref{fig:atlas_it_VC_appendix}, we show more reconstructed meshes in Atlasnet with the pretrained ${\rm{A}}{{\rm{t}}_{{\rm{CD}}}}$, our distance-oriented CD$^2$, ${\rm{CD}}_t^2$ and ${\rm{CD}}_p^2$.

\end{document}